\pdfoutput=1

\documentclass[journal,twoside,web]{ieeecolor}
\usepackage{tuffc}
\usepackage{cite}
\usepackage{amsmath,amssymb,amsfonts}
\usepackage{algorithmic}
\usepackage{graphicx}
\usepackage{textcomp}
\usepackage{wrapfig,colortbl}

\usepackage{ieee-tj-custom}

\newcommand{\cls}{\texttt{[CLS]}}

\newcommand{\res}[2]{\makecell{#1 \\ {\scriptsize $\pm$ #2}}}

\definecolor{abstractbg}{rgb}{1,0.969,0.914}
\def\BibTeX{{\rm B\kern-.05em{\sc i\kern-.025em b}\kern-.08em
    T\kern-.1667em\lower.7ex\hbox{E}\kern-.125emX}}
\begin{document}
\bstctlcite{IEEEexample:BSTcontrol} 
\title{Fusing Echocardiography Images and Medical Records for Continuous Patient Stratification}
\author{Nathan Painchaud, Jérémie Stym-Popper, Pierre-Yves Courand, Nicolas Thome, Pierre-Marc Jodoin, Nicolas Duchateau, and Olivier Bernard
\thanks{Manuscript received April 28, 2025; accepted August 16, 2025. This work was supported in part by the NSERC's Discovery Grants and Canada Graduate Scholarships-Doctoral programs, the FRQNT Doctoral Training Scholarships and the French National Research Agency (LABEX PRIMES [ANR-11-LABX-0063] of Université de Lyon, within the program "Investissements d'Avenir" [ANR-11-IDEX-0007], and the MIC-MAC [ANR-19-CE45-0005] and ORCHID [ANR-22-CE45-0029-01] projects). For the purpose of open access, the authors have applied a CC BY public copyright license to any Author Accepted Manuscript (AAM) version arising from this submission. }
\thanks{N. Painchaud, P.-Y. Courand, N. Duchateau, and O. Bernard are with Univ Lyon, INSA‐Lyon, Université Claude Bernard Lyon 1, UJM-Saint Etienne, CNRS, Inserm, CREATIS UMR 5220, U1294, F‐69621, Lyon, France (e-mail: nathan.painchaud@usherbrooke.ca). }
\thanks{N. Painchaud and P.-M. Jodoin are with the Department of Computer Science, University of Sherbrooke, Sherbrooke, QC, Canada. }
\thanks{J. Stym-Popper and N. Thome are with Sorbonne Université, CNRS, ISIR, F-75005, Paris, France. }
\thanks{P.-Y. Courand is also with the Cardiology Dept., Hôpital Croix-Rousse, Hospices Civils de Lyon, Lyon, France, and the Cardiology Dept., Hôpital Lyon Sud, Hospices Civils de Lyon, Lyon, France. }
\thanks{N. Thome and N. Duchateau are also with the Institut Universitaire de France (IUF). }}

\IEEEtitleabstractindextext{%
\fcolorbox{abstractbg}{abstractbg}{%
\begin{minipage}{\textwidth}\rightskip2em\leftskip\rightskip\bigskip
\begin{wrapfigure}[15]{r}{3in}%
\hspace{-3pc}\includegraphics[width=2.9in]{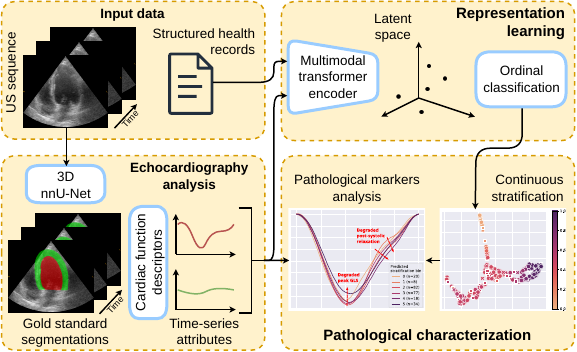}
\end{wrapfigure}%
\begin{abstract}
Deep learning enables automatic and robust extraction of cardiac function descriptors from echocardiographic sequences, such as ejection fraction or strain. These descriptors provide fine-grained information that physicians consider, in conjunction with more global variables from the clinical record, to assess patients' condition. Drawing on novel Transformer models applied to tabular data, we propose a method that considers all descriptors extracted from medical records and echocardiograms to learn the representation of a cardiovascular pathology with a difficult-to-characterize continuum, namely hypertension. Our method first projects each variable into its own representation space using modality-specific approaches. These standardized representations of multimodal data are then fed to a Transformer encoder, which learns to merge them into a comprehensive representation of the patient through the task of predicting a clinical rating. This stratification task is formulated as an ordinal classification to enforce a pathological continuum in the representation space. We observe the major trends along this continuum on a cohort of 239 hypertensive patients, providing unprecedented details in the description of hypertension's impact on various cardiac function descriptors. Our analysis shows that i) the XTab foundation model's architecture allows to reach high performance (96.8\% AUROC) even with limited data (less than 200 training samples), ii) stratification across the population is reproducible between trainings (within 5.7\% of mean absolute error), and iii) patterns emerge in descriptors, some of which align with established physiological knowledge about hypertension, while others could pave the way for a more comprehensive understanding of this pathology. Code is available at \url{https://github.com/creatis-myriad/didactic}.
\end{abstract}

\bigskip
\end{minipage}}}

\maketitle

\section{Introduction}
\label{sec:introduction}

\begin{table*}[!t]
\arrayrulecolor{subsectioncolor}
\setlength{\arrayrulewidth}{1pt}
{\sffamily\bfseries\begin{tabular}{lp{6.75in}}\hline
\rowcolor{abstractbg}\multicolumn{2}{l}{\color{subsectioncolor}{\itshape
Highlights}{\Huge\strut}}\\
\rowcolor{abstractbg}$\bullet$ & We proposed a framework to fuse structured EHR data with time-series descriptors derived from echocardiographic sequences, supervised by ordinal classification to represent the pathological continuum.\\
\rowcolor{abstractbg}$\bullet${\large\strut} & We compared multimodal strategies under limited data, illustrating the benefit of extracting relevant image descriptors and our method's major improvements over \sota\ (over 10 points in AUROC).\\
\rowcolor{abstractbg}$\bullet${\large\strut} & We applied our continuous stratification approach to a promising case study on hypertension, highlighting early warning patterns, both known and new, that mark the progression of hypertension.\\[2em]\hline
\end{tabular}}
\setlength{\arrayrulewidth}{0.4pt}
\arrayrulecolor{black}
\end{table*}

\IEEEPARstart{W}{hen} assessing patients' condition, physicians integrate relevant complementary data from various sources such as medical images and Electronic Health Records (EHRs)~\cite{pellegrini_unsupervised_2023} into a global picture of the patient's status. Such a clinical workflow is typical in cardiology for the characterization of hypertension (HT), a complex and multifaceted disease. While HT is the most prevalent cardiovascular disease (CVD), affecting around 1.28 billion adults worldwide~\cite{mancia_2023_2023}, it presents a complex pathophysiology involving multiple mechanisms and is associated with other risk factors of CVDs. Although the diagnosis for HT is based on widely recognized thresholds of measured blood pressure (BP), the risk of cardiovascular events related to HT is less clearly understood. Rule-based scores have been proposed to estimate these risks, but they are based on limited numbers of risk factors and require calibration to adjust to different populations~\cite{mancia_2023_2023}. For these reasons, there is clinical interest to develop models that can integrate multiple parameters of cardiac health to characterize precisely the HT continuum~\cite{zhang_breaking_2025}.

So far, most deep learning methods for cardiac applications have focused on highly-specific monomodal tasks, such as the automatic extraction of shape and motion/deformation parameters from MR and ultrasound images~\cite{sermesant_applications_2021,salte_artificial_2021,ling_extraction_2023}. Some methods combine detail-rich imaging data with other sources, but often sacrifice lots of information, for example by using slices from 3D volumes or frames from 2D+time sequences~\cite{hager_best_2023,schilcher_fusion_2024} or by resuming images to a few scalar biomarkers~\cite{zheng_pathological_2020,pellegrini_unsupervised_2023}. Furthermore, the most successful multimodal approaches leverage large datasets, while models trained on small pathology-specific datasets often lead to poor fitting~\cite{kline_multimodal_2022}. The difficulty of efficiently combining highly heterogeneous and limited data means that research on comprehensive models to aid higher-level diagnosis has lagged behind~\cite{rajpurkar_ai_2022}.

Recently, the Transformer architecture has shown promising results in multimodal applications, combining different data sources naturally since it makes no assumptions about their structure~\cite{xu_multimodal_2023}. It is also competitive in monomodal tasks~\cite{xu_multimodal_2023}, for example breaking through on tabular data where deep learning based approaches previously struggled against simpler machine learning solutions~\cite{gorishniy_revisiting_2021}. However, in practice, the lack of data priors means that Transformers are trained on larger datasets to learn the structure in the data for themselves~\cite{xu_multimodal_2023,zhu_xtab_2023}. While this adaptability is at the core of their success, it exacerbates issues with limited volumes of data on which to train, especially in healthcare. The recent introduction of foundation models could unlock the potential of Transformers for medical applications. These are models pretrained on enormous quantities of varied data, assuming they can efficiently adapt to downstream tasks with limited quantities of domain-specific data~\cite{zhang_challenges_2024}.

In this paper, we use Transformers to combine a small dataset of EHR data and cardiac function descriptors extracted from 2D+time echocardiographic sequences to stratify HT. Our method is built around the tabular representation of EHR data, with a dedicated branch to integrate image-based data. We also use the stratification labels' order as additional supervision to learn a more informative representation of patients and predict a position along a pathological continuum~\cite{zhang_breaking_2025}. Finally, we showcase how this continuous representation can help retrospectively discover new subtle indicators of hypertension's progress on real-world, targeted clinical cohorts.

\section{Related Work}
\label{sec:related}

\subsection{Multimodal Machine Learning}
\label{sec:related:multimodal}
Multimodal learning encompasses multiple techniques for combining heterogeneous data for downstream tasks. Our application aims to fuse complementary EHR and imaging data into a joint representation from which to make predictions~\cite{baltrusaitis_multimodal_2019}. In this setting, mid-fusion approaches have shown promising results~\cite{nagrani_attention_2021,zhou_transformer-based_2023}. They consist in having intermediate neural network layers independently extract features from each modality, before the features are joined and processed by further layers. By contrast, in early fusion, feature extractor components are frozen and cannot learn features adapted to the fusion task~\cite{huang_fusion_2020}. Despite mid-fusion's adaptability, early fusion is still favored in the biomedical community~\cite{huang_fusion_2020,kline_multimodal_2022}, since it can be achieved by simply joining features extracted using pretrained models~\cite{huang_fusion_2020}.

As mentioned in the introduction, research has also highlighted the intrinsic advantages of Transformers for multimodal applications~\cite{xu_multimodal_2023}. Their lack of assumptions about the structure of the data allows them to easily adapt to different modalities, compared to the spatial structure bias in convolutional networks.

Most multimodal research has focused on modalities that are cheap to acquire and abundant, e.g. images, text and audio~\cite{baltrusaitis_multimodal_2019}. However, modalities rarer elsewhere are common in healthcare data, e.g. structured data, which motivated frameworks designed specifically for medical tabular and imaging data. Thus, Hager \etal~\cite{hager_best_2023} and Schilcher \etal~\cite{schilcher_fusion_2024}  introduced multimodal frameworks tailored to medical tabular and imaging data. MultiModal Contrastive Learning (MMCL), proposed by Hager \etal\ aligns modalities during training instead of fusing them, to perform monomodal inference on images. As for Schilcher \etal, they implemented multiple fusion schemes (early, mid, and late) which brought slight but systematic improvements over unimodal image predictions. Both used large datasets, with over 40K subjects in the case of MMCL and 1\,073 for Schilcher \etal. Finally, Zhou \etal\ proposed IRENE~\cite{zhou_transformer-based_2023}, a generic multimodal Transformer for disease diagnosis with a custom cross-attention component between images and unstructured textual EHR data. IRENE was trained and evaluated separately on two large datasets of 2\,362 and 44K+ subjects. Considering the scales of data the above methods were trained on even though they target healthcare data, they are likely to run into data scarcity issues on our application.

\subsection{Patient Stratification}
\label{sec:related:stratification}
Stratification refers to the prediction of the outcome or pathological stage of patients~\cite{schlesinger_deep_2020}. Like multimodal methods, stratification models often rely on large quantities of unstructured data to unveil patterns, e.g. EHRs of 1.6M+ patients used in \cite{landi_deep_2020}. However, models with such broad scopes risk under-performing on rare subtypes~\cite{oakden-rayner_hidden_2020}. This shortcoming calls for targeted methods that can leverage as much information as possible from scarce data.

In cardiology, deep learning models have mostly focused on quantifying imaging data over stratifying patient risk~\cite{schlesinger_deep_2020,sanchez-martinez_machine_2022}. Still, some works used representation learning and clustering to perform unsupervised stratification~\cite{landi_deep_2020,loncaric_automated_2021}. The different properties of supervised and unsupervised stratification methods lead to a dilemma. Unsupervised methods can learn rich and continuous representations, which highlight gradual alterations in biomarkers~\cite{loncaric_automated_2021}. On the downside, these representations might not align with established clinical scales. Supervised approaches are anchored to such scales~\cite{schlesinger_deep_2020}, but low-density regions in their clustered representations can fail to characterize full pathological continuums.

However, supervised methods typically ignore the ordering of stratification labels, e.g. from best to worst outcome, which could help structure their representation. Stratification can be framed as an ordinal classification problem, in which models' predictions should take into account the ordering of the labels. Until recently, such methods implied more complex models and/or optimizations, and practical implementations were limited to linear models~\cite{tran_stabilized_2015}. Now, ordinal classification has been adapted for deep learning in a simple formulation applicable on top of any feature-extractor architecture~\cite{beckham_unimodal_2017}.

\section{Method}
\label{sec:method}

\begin{figure*}[thb]
    \includegraphics[width=\textwidth]{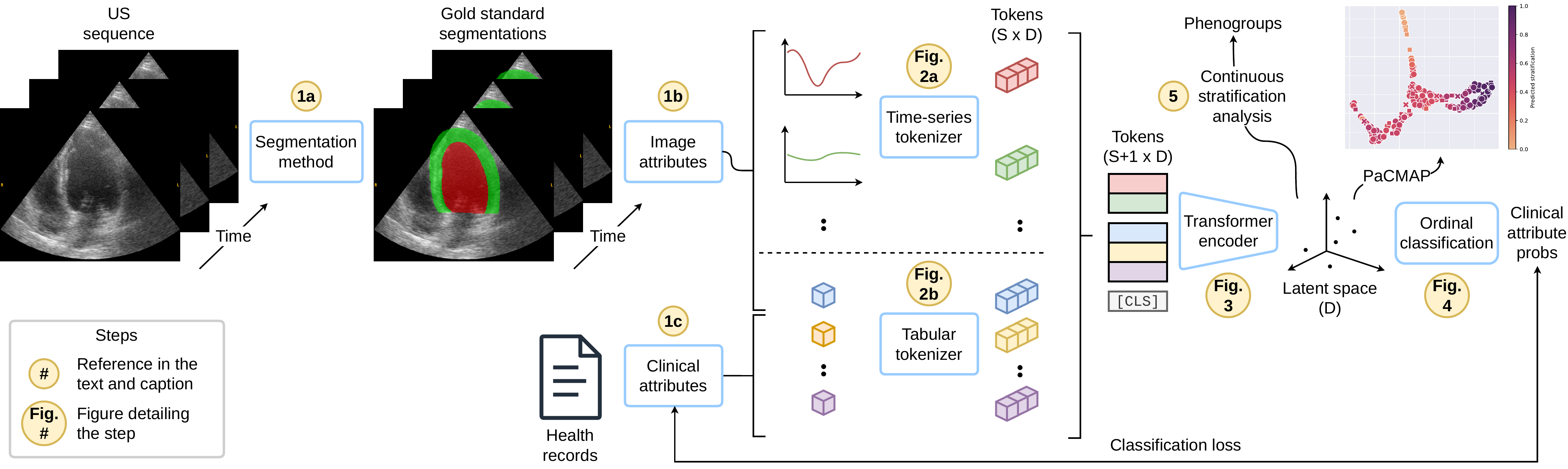}
    \caption{Schema of our multimodal fusion pipeline. First, echocardiograms are segmented (1a) and descriptors are extracted from them (1b). In parallel, health records data is structured into categorical and scalar descriptors (1c). Time-series descriptors (with respect to the cardiac cycle) and tabular descriptors are processed into individual embeddings using modality-specific tokenizers (\cref{fig:time-series_tokenizer}, \cref{fig:tabular_tokenizer}). The embeddings are then fed to a Transformer encoder (\cref{fig:transformer-encoder}) trained to predict a representation which takes into account the ordering of the target classes (\cref{fig:ordinal_classification}). The learned representation can finally be analyzed according to the predicted stratification (5) or visualized using dimensionality reduction (PaCMAP~\cite{wang_understanding_2021}).}
    \label{fig:method}
\end{figure*}

\begin{figure*}[htb]
    \begin{subfigure}[b]{0.56\textwidth}
        \includegraphics[width=\textwidth]{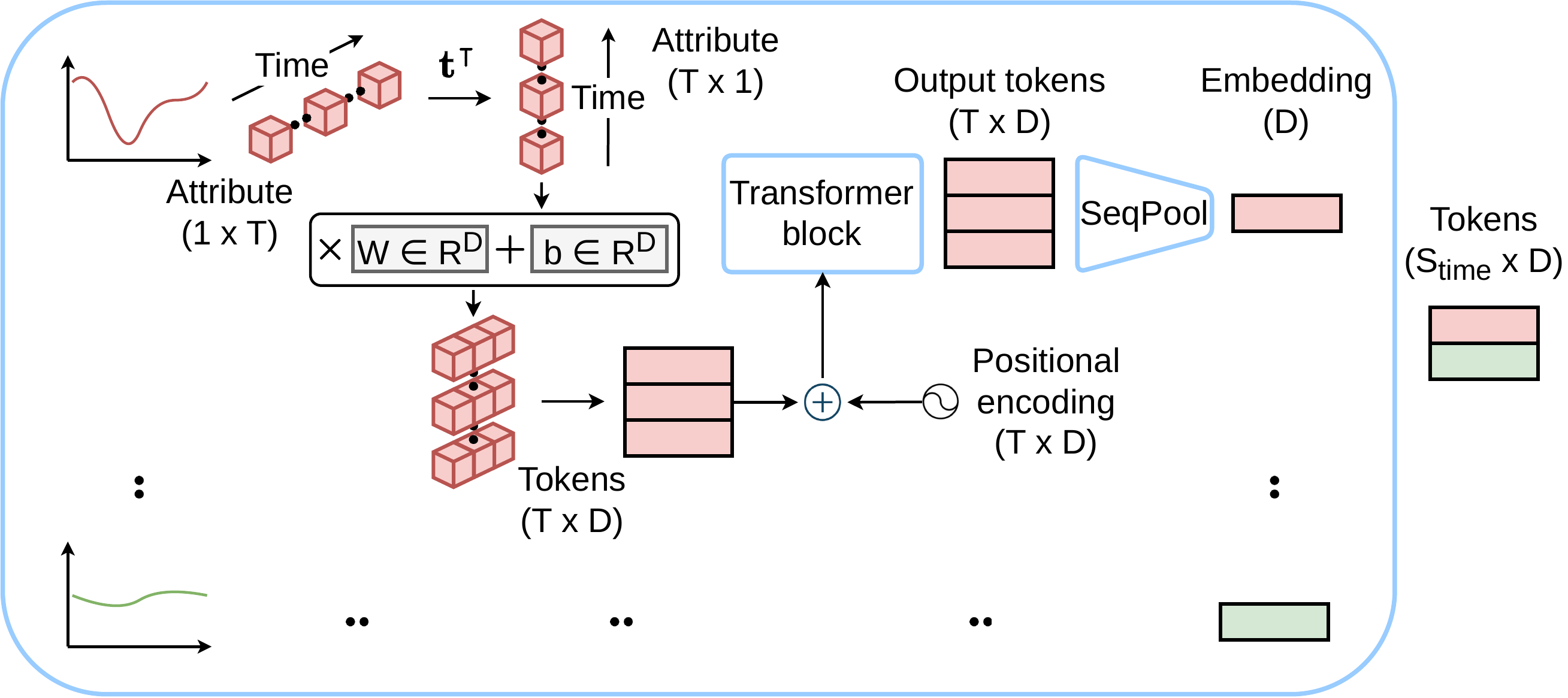}
        \caption{Time-series Tokenizer (cf. \cref{sec:method:tokenization:time-series}): Each time-series $\var{t}$ is expanded into a sequence of tokens by projecting each step's value to $D$ dimensions. Sequences are then processed independently by the same Transformer, to obtain an output embedding, i.e. token, for each time-series.}
        \label{fig:time-series_tokenizer}
    \end{subfigure}
    \hfill
    \begin{subfigure}[b]{.39\textwidth}
        \includegraphics[width=\textwidth]{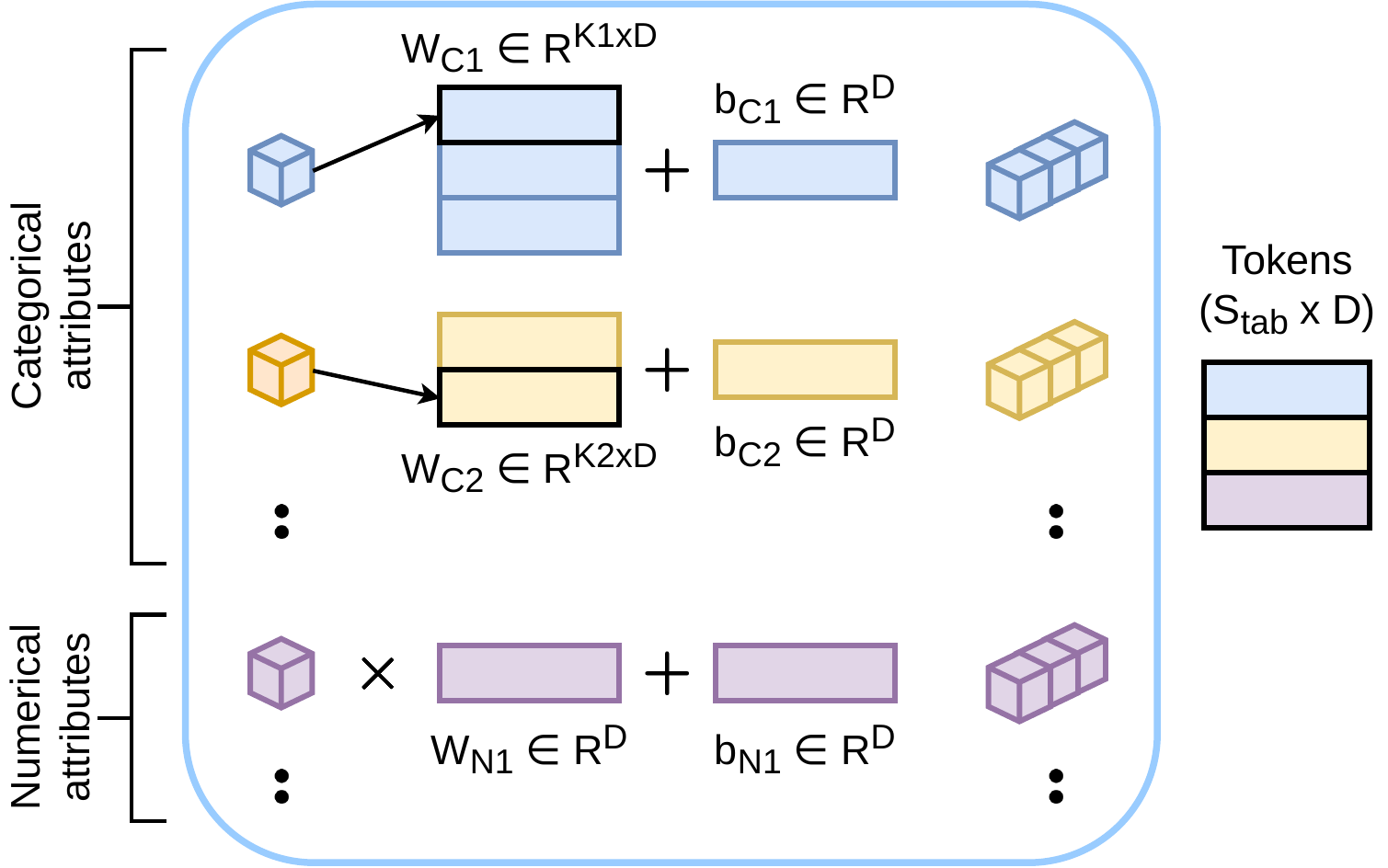}
        \caption{Tabular Tokenizer (cf. \cref{sec:method:tokenization:tabular}): Following \cite{gorishniy_revisiting_2021}, learnable layers, dependent on datum type, embed each tabular descriptor as a $D$-dimensional token. Categorical data [top] use dictionaries of embeddings, and scalar data [bottom] use linear layers.
        }
        \label{fig:tabular_tokenizer}
    \end{subfigure}
    \caption{Details of the two modality-specific tokenization methods for embedding multimodal data. The output of each method is a sequence of tokens of length equal to the number of input descriptors for each modality: $S_{\text{time}}$ and $S_{\text{tab}}$ for time-series and tabular descriptors, respectively.}
    \label{fig:tokenizers}
\end{figure*}

\Cref{fig:method} summarizes our pipeline. The echocardiograms are first segmented (1a) and numerical and time-series descriptors are extracted (1b). For health records data, we manually extracted categorical (e.g. sex, medical history, etc.) and numerical descriptors (e.g. age, BMI, etc.) from EHRs and a posterior assessment by a cardiologist (1c). Each modality is processed independently (\cref{fig:time-series_tokenizer,fig:tabular_tokenizer}) before being fed to an encoder (\cref{fig:transformer-encoder}) that performs data fusion and outputs a joint embedding for downstream tasks.

Since we are interested in learning a meaningful latent representation at the output of the encoder, we use a specific formulation of the supervised training objective in our pipeline. We frame the prediction of the stratification labels as an ordinal classification, where continuous values regressed by the model are converted to discrete labels~\cite{beckham_unimodal_2017}, to integrate the stratification notion in the latent representation. With this, we obtain a continuous value between 0 (healthy) and 1 (severe disease) for each patient, indicating the model's prediction along the pathological continuum. This formulation allows for an efficient traversal of the manifold without having to resort to ad hoc methods.

Below, we describe steps 2 through 4 illustrated in \cref{fig:method}. Since steps 1 and 6 are dataset or application-specific, they are further explained in \cref{sec:experimental_setup:data,sec:experimental_setup:image,sec:results:continuous_strat}, respectively.

\subsection{Multimodal Tokenization (Step 2)}
\label{sec:method:tokenization}
From the application's perspective, our framework operates on images and EHR data. As mentioned above, these raw data are preprocessed to extract i) time-series descriptors from images (series of scalar values per patient), and ii) tabular descriptors from EHRs and images (global categorical and numerical variables per patient). Given the fundamentally different structures of these data types, we split the tokenization in two streams ($S_{\text{time}}$ and $S_{\text{tab}}$), one for each modality (cf. \cref{fig:tokenizers}).

\subsubsection{Time-Series Tokenizer}
\label{sec:method:tokenization:time-series}
Under the assumption that there are (possibly complex) patterns in time-series descriptors during the cardiac cycle, they require modality-specific processing to be properly extracted. Therefore, we took inspiration from how Pellegrini \etal\ integrated longitudinal data in their multimodal Transformer~\cite{pellegrini_unsupervised_2023}
for the design of our time-series tokenizer module, detailed in \cref{fig:time-series_tokenizer}.

Starting with a vectorized input sequence $\var{t} \in \mathbb{R}^T$, each scalar in $\var{t}$ is upscaled to the embedding size $D$ through a linear layer. This $D$-fold increase in dimensionality is not meant to transform the data, but rather to structure it so that downstream attention operations can more meaningfully process it. Thus, the sequence, now made up of $T$ tokens of size $D$, is combined to a positional encoding before being fed to a Transformer block. Finally, the $T$ output tokens are reduced to a single $D$-dimensional token --- the embedding for the time-series --- using sequence pooling, i.e. a dynamic weighted averaging~\cite{hassani_escaping_2022}. Since all the time-series descriptors represent the same type of data, i.e. cyclical 1D signals, the weights of their tokenizer are shared between time-steps and descriptors.

\subsubsection{Tabular Tokenizer}
\label{sec:method:tokenization:tabular}
The recent FT-Transformer framework by Gorishniy \etal~\cite{gorishniy_revisiting_2021} achieved \sota\ results on various tasks involving tabular data. Therefore, we adopted their Feature Tokenizer for tabular data, illustrated in \cref{fig:tabular_tokenizer}.

Tabular data is represented so that each descriptor/column maps to a token. This means that each patient/row corresponds to a sequence of tokens that can be processed by a Transformer. Thus, tokenization upsamples each descriptor to a $D$-dimensional token. For categorical descriptors, the upsampling operation is an embedding layer, i.e. a dictionary of embedding vectors for each label to add to a shared bias. For numerical descriptors, it is a linear layer. To learn distinct embeddings for each descriptor, the parameters are not shared between descriptors.

\subsection{Transformer Encoder (Step 3)}
\label{sec:method:transformer_encoder}
\begin{figure}[tbh]
    \includegraphics[width=\columnwidth]{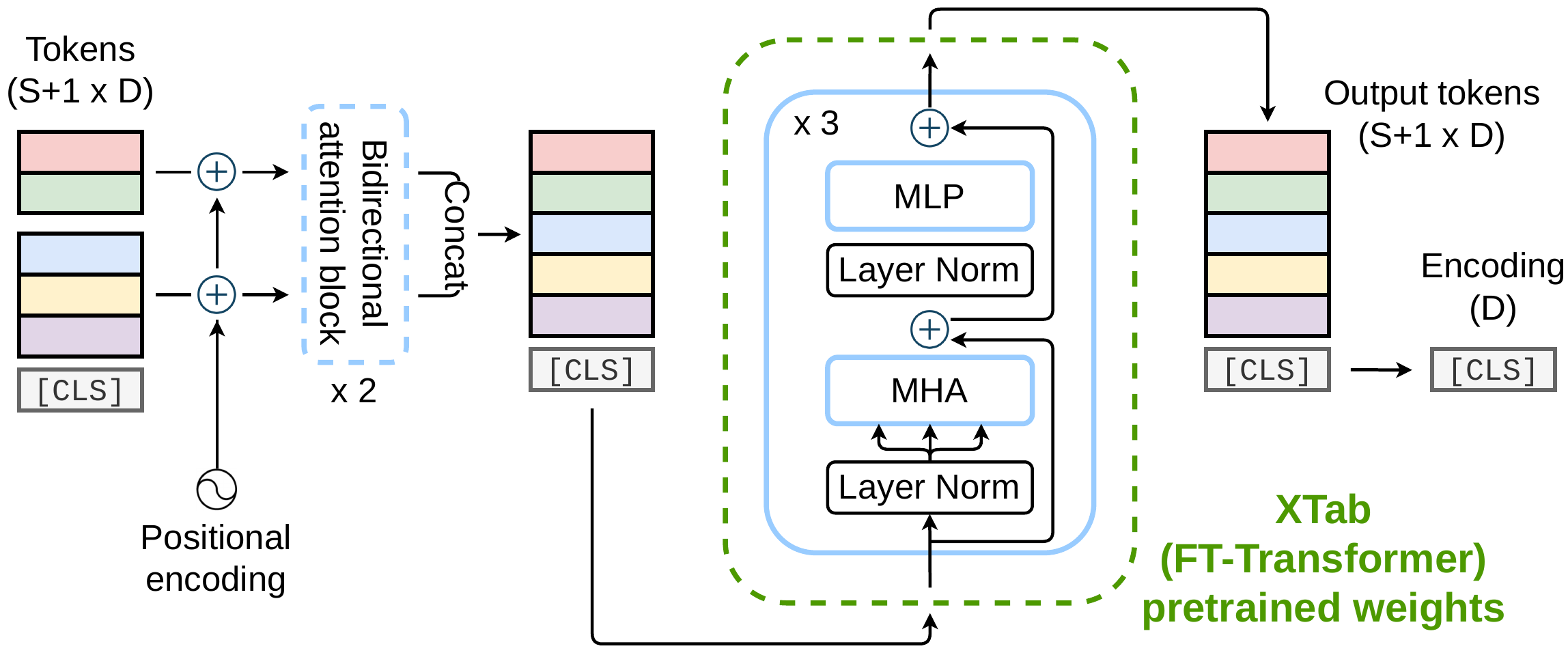}
    \caption{Detailed architecture of the Transformer encoder block from \cref{fig:method}. The tokens, first separated by modality, optionally pass through bidirectional multimodal attention blocks~\cite{zhou_transformer-based_2023}, before being concatenated into one unified sequence of tokens. They are then fed to self-attention Transformer blocks. In experiments using the XTab foundation model's configuration~\cite{zhu_xtab_2023}, these self-attention blocks (green dashes in the fig.) may be initialized with XTab's pretrained weights. In all configurations, other components, such as the \cls\ token, positional encoding, and tokenizers, are initialized randomly.}
    \label{fig:transformer-encoder}
\end{figure}

The tokenization described in \cref{sec:method:tokenization} outputs two sequences of tokens (one per modality). Our pipeline concatenates both sequences and a \cls\ token to get one sequence of $S+1$ tokens, where $S = S_{\text{time}} + S_{\text{tab}}$, which is fed to a Transformer encoder,  detailed in \cref{fig:transformer-encoder}.

This combination of Transformer backbone and data representation grants flexibility to the model. Since input descriptors map to tokens, and Transformers handle token sequences of various lengths, the same backbone can be re-used across different configurations of input data.

\subsubsection{Multimodal Token Fusion}
\label{sec:method:transformer_encoder:token_fusion}
Our recommended pipeline simply concatenates tokens from different modalities before the self-attention Transformer encoder. As an alternative, we tested a \sota\ multimodal Transformer module. It consists of bidirectional multimodal attention blocks introduced in the IRENE model~\cite{zhou_transformer-based_2023}, which performs symmetric cross-attention on both modalities before concatenating tokens. This optional component was added as shown in \cref{fig:transformer-encoder}, and its impact is discussed in \cref{sec:results:multimodal_fusion:tokenization}.

\subsubsection{FT-Transformer \& XTab Foundation Model}
\label{sec:method:transformer_encoder:foundation_model}
Once time-series and tabular tokens are concatenated, we use a vanilla Transformer encoder, like in the FT-Transformer framework~\cite{gorishniy_revisiting_2021}. The configuration --- number of layers, token size, normalization, etc. --- was chosen to follow the one used by XTab, a tabular foundation model built on a FT-Transformer backbone~\cite{zhu_xtab_2023}.

XTab's approach to generalization differs from that of other foundation models in domains such as natural language processing (NLP) and CV~\cite{zhu_xtab_2023}. Since the meaning and context of tabular data varies between datasets, XTab foregoes learning a universal tokenizer, focusing instead on learning an initialization of weights that generalizes well to downstream tasks. XTab's pretrained weights only include the Transformer blocks, indicated by the green box in \cref{fig:transformer-encoder}, since they do not depend on the types and number of input features. Other parameters specific to the task or data, such as tokenizers or the \cls\ token, are learned from scratch.

To initialize its weights, XTab is trained, in a self-supervised manner, to reconstruct features from 52 tables from the AutoML benchmark (AMLB)~\cite{gijsbers_amlb_2023}. Across tables, the size of the data varies greatly in terms of numbers of rows (between 100 and 10\,000\,000) and columns (between 4 and 14\,892).

\subsection{Ordinal Classification (Step 4)}
\label{sec:method:ordinal_classification}

\begin{figure}[t]
    \includegraphics[width=\columnwidth]{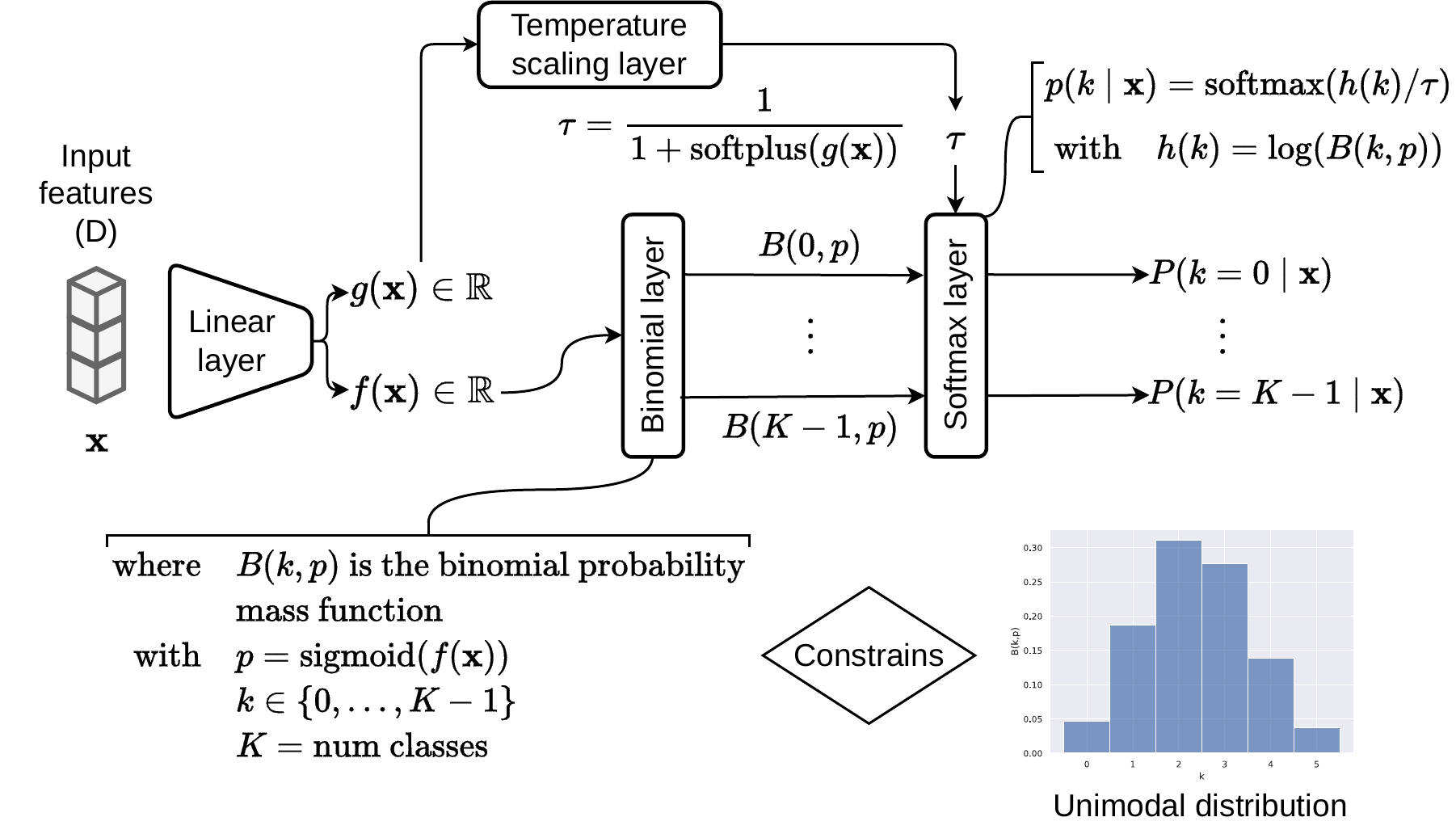}
    \caption{Details of the ordinal classification from \cref{fig:method} (\ref{fig:ordinal_classification}). Given a feature vector $\var{x}$, the goal is to output probabilities $p(k \mid \var{x})$ over the $K$ classes following a unimodal distribution. This means probabilities should gradually decrease on both sides of the predicted class. To achieve this, the model predicts two values, $f(x)$ and $g(x)$, instead of logits. The success probability $p \in [0,1]$ of a binomial is derived from $f(x)$, and its probability mass function $B(k,p)$ is used to compute per-class probabilities, ensuring a unimodal distribution. Finally, a softmax is applied on the probabilities, with a temperature derived from $g(x)$. In the end, the probabilities $p(k \mid \var{x})$ can be used with standard classification losses.}
    \label{fig:ordinal_classification}
\end{figure}

One goal of our method is to provide a continuous stratification of HT severity. To achieve this, we use the implicit information in the ordering of discrete labels provided by an expert cardiologist. Our method is constrained to learn a continuous representation along which patients are ordered with respect to (w.r.t.) HT severity labels by relying on the ordinal classification formulation proposed by Beckham and Pal~\cite{beckham_unimodal_2017}, detailed in \cref{fig:ordinal_classification}. Given the features extracted by the encoder, a linear prediction head predicts the probability of success $p \in [0,1]$ of a binomial distribution. The logits are computed analytically using $p$ and the binomial probability mass function. This allows the constrained prediction head to be used with standard classification losses, like cross-entropy.

\begin{figure}[thb]
    \begin{subfigure}[t]{\columnwidth}
        \begin{center}
        \includegraphics[width=0.9\textwidth]{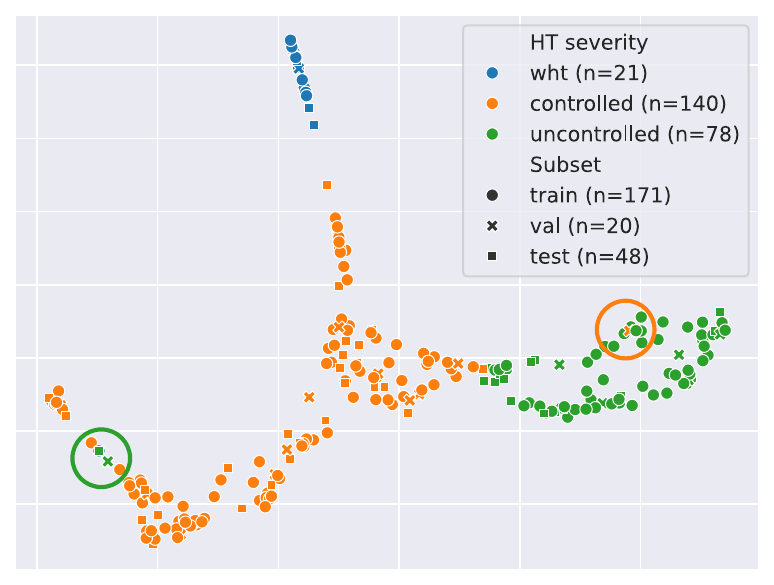}
        \end{center}
        \caption{Embedding colored w.r.t. the target degrees of hypertension, ordered in the legend by their severity (cf. \cref{sec:experimental_setup:data}). Misclassified patients appear outside of their respective regions and are marked by circles colored like their target labels.}
        \label{fig:target_embedding}
    \end{subfigure}
    \begin{subfigure}[t]{\columnwidth}
        \begin{center}
        \includegraphics[width=0.9\textwidth]{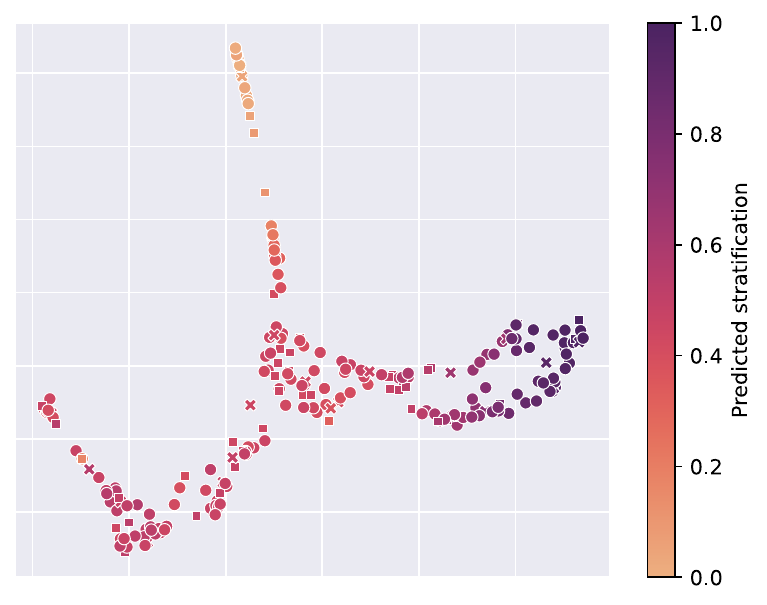}
        \end{center}
        \caption{Embedding colored w.r.t. the continuous stratification predicted by our method, where 0 (light) corresponds to healthy subjects, and 1 (dark) corresponds to the most hypertensive patients.}
        \label{fig:continuum_embedding}
    \end{subfigure}
    \caption{2D visualizations, using the PaCMAP dimensionality reduction algorithm~\cite{wang_understanding_2021}, of the 192D latent space learned by our Transformer encoder. Points correspond to patients, colors are linked to attributes of interest, and markers represent the dataset partition used by the model.}
    \label{fig:embeddings}
\end{figure}

This formulation means $p$ determines the label with the most probability mass, and reducing/increasing $p$ shifts the probability mass center towards earlier/later labels in the ordering, respectively. Therefore, $p$ can be used directly as a continuous stratification, rather than converted to probabilities over the labels. \Cref{fig:embeddings} contrasts an embedding of the patients colored w.r.t. discrete labels \subref{fig:target_embedding} and $p$ \subref{fig:continuum_embedding}, illustrating how our method manages to learn a continuous representation structured around the labels. Since the prediction head is linear, it forces a direction along the latent representation to correspond to the order of target labels, clearly visible in \cref{fig:target_embedding}.

\section{Experimental Setup}
\label{sec:experimental_setup}

\subsection{Dataset}
\label{sec:experimental_setup:data}
To evaluate our pipeline, we used an in-house dataset of echocardiographic sequences and EHR data collected from 239 hypertensive patients (CARDINAL cohort). According to French law, this retrospective study was approved by the local ethics committee (Scientific and Ethical Committee of Hospices Civils de Lyon, France, CSE-HCL – IRB\,0013204, 13/09/2022, number 22\_598, MR-004). In accordance with legislation in France at the time of the study, an information letter was sent to all participants who were still alive, or their next of kin if they were not, to give them the opportunity to refuse the use of their data for this study. The CARDINAL imaging data was first introduced by Ling \etal\ to train a 2D+time nnU-Net using pseudo-labels~\cite{ling_extraction_2023}.

For the current study, we included an additional 62 numerical and categorical descriptors, extracted from an EHR server or derived from other descriptors. The descriptors cover various aspects of the patients' condition, from general information (e.g. age, sex, etc.), medical history (e.g. stroke, tobacco, etc.), and biological reports (e.g. NT-proBNP, etc.) to measures from hospital stays (e.g. 24-hour ambulatory blood pressure) and transthoracic echocardiograms (e.g. Apical 4 Chambers (A4C), etc.). All the descriptors are described in tables VI to VIII in the supplementary materials.

We used the HT severity descriptor as prediction target. It represents three degrees of hypertension assessed post hoc by an expert cardiologist. The labels, in ascending order of severity~\cite{franklin_white-coat_2013}, are: i) \textit{wht} (White coat HyperTension, n=21), meaning false positive suspicions of hypertension, ii) \textit{controlled} (n=140), for patients whose hypertension is under the recommended blood pressure level given their treatment, and iii) \textit{uncontrolled} (n=78), for patients who remain above the recommended blood pressure level despite treatment. Finally, we split the dataset using a 5-fold cross-validation strategy, with a 70/10/20 ratio between training, validation, and test sets in each split.

\subsection{Echocardiograms Processing}
\label{sec:experimental_setup:image}
As described in \cref{sec:method}, the 2D+time echocardiographic sequences are first segmented, from which global and frame-by-frame cardiac function descriptors are automatically computed. For the segmentations, we use the \sota\ masks predicted by the nnU-Net from Ling \etal~\cite{ling_extraction_2023} as gold-standard, with automatic quality insurance for anatomical~\cite{painchaud_cardiac_2020} and temporal~\cite{painchaud_echocardiography_2022} coherence. This nnU-Net reports left ventricle segmentations across full cardiac cycles with a Dice of 96.9, a Hausdorff distance (HD) of 2.3 mm and a mean absolute distance (HD) of 0.7 mm. These results are within the intra-observer variability previously reported for echocardiographic LV segmentation on the CAMUS dataset, of 93.8 for Dice, 4.6 mm for HD, and 1.4 mm for MAD~\cite{leclerc_deep_2019}, meaning that the model is as reliable as clinicians.

The descriptors derived from the segmentations were chosen by a cardiologist based on their relevance for assessing HT. The scalar descriptors measured on the patients are the following:
\begin{itemize}
    \item (\textit{2 descriptors}) Left Ventricle (LV) volumes at End-Diastole (ED) and End-Systole (ES), using Simpson's biplane method~\cite{folland_assessment_1979};
    \item (\textit{1}) Ejection Fraction (EF) from LV ED/ES;
    \item (\textit{8}) Mininum/maximum myocardial curvature of the walls (A4C: septal/lateral, A2C: inferior/anterior) at ED~\cite{marciniak_septal_2021}, in both A4C and A2C views.
\end{itemize}

The time-series descriptors measured on both A4C and A2C views are:
\begin{itemize}
    \item (\textit{2 \texttimes\ 1 descriptor}) Surface of the LV;
    \item (\textit{2 \texttimes\ 1}) LV length along its main axis (base to apex);
    \item (\textit{2 \texttimes\ 3}) Global strain and regional strain, with the local segments depending on the view (A4C: septal/lateral, A2C: inferior/anterior);
    \item (\textit{2 \texttimes\ 2}) Local average myocardial thickness, with local segments depending on the view (A4C: septal/lateral, A2C: inferior/anterior).
\end{itemize}

We define the septal/lateral and inferior/anterior segments as the first 30\% of the myocardium from base to apex on each side, just as in~\cite{marciniak_septal_2021}. Because we are not doing tracking, we split the myocardium into longitudinal segments by sampling equidistant control points along the myocardium and selecting a subset of control points.

\subsection{Framework Configuration}
\label{sec:experimental_setup:framework_config}

\begin{table}[t]
\centering
\caption{Details of tested FT-Transformer configurations.}
\label{tab:fttransformer_configurations}
\begin{NiceTabular}{l cc}
\toprule
\Block{2-1}{Hyperparameter} & \Block{1-2}{Model configuration} & \\
\cmidrule(lr){2-3}
    & \makecell{Tiny \\ FT-Transformer} & \makecell{XTab~\cite{zhu_xtab_2023} \\ (Large FT-Transformer)} \\
\midrule
Embedding size & 8 & 192 \\
Nb Transformer blocks & 6 & 3 \\
Nb attention heads & 2 & 8 \\
\midrule
Optimizer & \Block{1-2}{AdamW} & \\
Optimizer params & \makecell{learning rate \\ finder} & \makecell{learning rate: 1e-4 \\ L2 regularization: 1e-5} \\
Pretrain/finetune steps & 2\,500 / 500 & -- / 2\,000 \\
Rand. init. train steps & 2\,500 & 2\,000 \\
\bottomrule
\end{NiceTabular}
\end{table}

We tested two main configurations for the architecture of the Transformer encoder, detailed in \cref{tab:fttransformer_configurations}. XTab is the exact configuration proposed in~\cite{zhu_xtab_2023}. Given XTab's large size (\textasciitilde1.1M params) compared to our training samples (<200), we performed a grid search over hyperparameters values leading to smaller models to propose an optimal ``tiny'' model (\textasciitilde5K params).

We also performed ablation studies over other aspects of our pipeline, described below and discussed later in \cref{sec:results:multimodal_fusion}.

\begin{itemize}
    \item \emph{Input data}: Given our wealth of tabular data, we studied the impact of the presence or absence of certain descriptors. We defined subsets of tabular descriptors to provide as inputs to models. \textit{All} refers to all 62 tabular descriptors. Many of these descriptors come from ultrasound exams, e.g. measures of cardiac structures performed by experts on echocardiograms. Leaving out these descriptors, we obtained 28 \textit{clinical only} descriptors;
    \item \emph{Image descriptors tokenization \& fusion}: We stated in \cref{sec:method:tokenization:time-series} that time-series descriptors require modality-specific processing. To validate this, we compared our time-series Transformer tokenizer to an approach that linearly projects time-series to the tokens’ $D$-dimensional embedding. Concurrently, we tested adding bidirectional multimodal attention blocks, given their reported success on medical data~\cite{zhou_transformer-based_2023}, to see if first mixing modalities' information through cross-attention could offset the lack of modality-specific processing;
    \item \emph{Latent representation}: We investigated the optimal way to extract a representation from Transformer features, between using \cls\ token and performing a weighted average of all output tokens using sequence pooling. At the same time, we tested the impact of the \textit{ordinal constraint}, described in \cref{sec:method:ordinal_classification}, compared to the classification head proposed by the FT-Transformer~\cite{gorishniy_revisiting_2021} with a cross-entropy loss.
\end{itemize}

\subsection{State-of-the-art Comparisons}
\label{sec:experimental_setup:sota}
Aside from ablation studies of our proposed framework, we compare it to two relevant \sota\ methods, first mentioned in \cref{sec:related:multimodal}. The purposes of comparing to these methods are summarized below. We also explain how they were adapted to work with our data. The results of all three methods are discussed in \cref{sec:results:multimodal_fusion}.

\subsubsection{Image-centered Framework}
\label{sec:experimental_setup:sota:bob}
To assess the benefit of extracting clinically relevant time-series descriptors from images, we tested the multimodal pipeline proposed by Hager \etal~\cite{hager_best_2023}, which works with images directly. Their model, MMCL, encodes images and tabular data separately using a ResNet50 and MLP, respectively, before aligning both sets of features using a contrastive strategy. Classification is performed by a linear layer on top of the image features. At test-time, only the image branch and linear classifier are used.

This method is meant for 2D images, with the authors selecting important slices when dealing with 3D images. To be fair in our comparison and provide temporal information, we concatenate the 4 most important frames of each patient (i.e. End-Diastole (ED) and End-Systole (ES) frames from both A4C and A2C views) as different channels to the ResNet50.

\subsubsection{Multimodal Transformer}
\label{sec:experimental_setup:sota:irene}
We also tested IRENE~\cite{zhou_transformer-based_2023}, a \sota\ multimodal Transformer model for healthcare data. IRENE was originally tested on images and a combination of structured and unstructured clinical data. They first process tokens with blocks mixing cross-attention and self-attention modules, called \textit{bidirectional multimodal attention}, before using standard self-attention blocks on the concatenated tokens from both modalities. Even though IRENE was tested on data of a different nature than our application, its Transformer backbone allowed us to swap in our time-series and tabular data tokens, to compare only their multimodal fusion encoder to ours.

\subsection{Computational Resources}
\label{sec:experimental_setup:computation}
Most of the computational cost comes from segmenting echocardiographic sequences ($\sim$40s per patient) and computing the time-series descriptors ($\sim$10s per patient) (cf. \cref{sec:experimental_setup:image}). In contrast, the computation of the multimodal fusion model is negligible (<1ms per patient). The times reported above were measured on a computer with an Intel Xeon W-2255 CPU (used to compute time-series descriptors) and a 16GiB Quadro RTX 5000 GPU (used by the segmentation and multimodal fusion models).

\section{Results}
\label{sec:results}

\subsection{Multimodal Fusion Framework}
\label{sec:results:multimodal_fusion}
\Cref{tab:ablation_data_architecture_init,tab:sota_comparison,tab:ablation_multimodal_fusion,tab:ablation_latent_representation} present the results of the ablation studies on our proposed pipeline described in \cref{sec:experimental_setup:framework_config}, and of the comparisons with \sota\ multimodal methods for imaging and clinical data described in \cref{sec:experimental_setup:sota}. All tables report the area under the receiver operating characteristic curve (AUROC) for the HT severity labels (cf. \cref{sec:experimental_setup:data}), using a one-vs-rest scheme given the multi-class setting. \Cref{tab:ablation_multimodal_fusion,tab:ablation_latent_representation}, which describe ablation studies on our proposed pipeline’s components, also report per-class accuracies. In the following subsections, we analyze how each design choice affects the performance, and the properties of our framework highlighted by these changes. It is important to note that classification performance is not the ultimate goal of this study, but rather a target to help structure the latent space. Indeed, we hypothesize that the better the classification scores, the more relevant the underlying representation is for characterizing HT. This assumption is justified in more details at the beginning of \cref{sec:results:continuous_strat}.

\begin{table*}[t]
\centering
\caption{Ablation study of the descriptors provided as input, and of the Transformer encoder's configuration and weights initialization. Results correspond to the AUROC's mean ± standard deviation over the test sets in a 5-fold cross-validation. Configurations statistically different from the best configuration (p<0.05 using a one-sided Wilcoxon signed-rank test) are marked with an asterisk (*).}
\label{tab:ablation_data_architecture_init}
\begin{NiceTabular}{ccc cc ccc}
\toprule
\Block{2-1}{\makecell{Tabular \\ descriptors}} & \Block{2-1}{\makecell{Time-series \\ descriptors}} & \Block{2-1}{\makecell{\# \\ descriptors}} & \Block{1-2}{Tiny FT-Transformer [cf. \cref{tab:fttransformer_configurations}]} & & \Block{1-3}{XTab (Large FT-Transformer) \cite{zhu_xtab_2023}} \\
\cmidrule(lr){4-5} \cmidrule(lr){6-8}
    & & & \makecell{random \\ weights init.} & \makecell{\# \\ parameters} & \makecell{random \\ weights init.} & \makecell{XTab \\ weights init.} & \makecell{\# \\ parameters} \\
\cmidrule(lr){1-3}
    \cmidrule(lr){4-5}
    \cmidrule(lr){6-8}
\Block{2-1}{all} & \xmark & 62 & 91.1 $\pm$ 6.7 * & 5.8K & 95.6 $\pm$ 5.5 & 89.8 $\pm$ 6.0 * & 898K \\
    & \cmark & 76 & 89.3 $\pm$ 7.3 * & 6.9K & 95.3 $\pm$ 3.1 & 88.2 $\pm$ 4.9 * & 1\,137K \\
\cmidrule(lr){1-8}
\Block{2-1}{\makecell{clinical \\ only}} & \xmark & 28 & 92.1 $\pm$ 4.3 * & 4.9K & \underline{95.8 $\pm$ 3.2} & 90.3 $\pm$ 6.4 * & 876K \\
    & \cmark & 42 & 92.7 $\pm$ 5.5 * & 6.0K & \textbf{96.8 $\pm$ 3.1} & 93.8 $\pm$ 3.8 *& 1\,115K \\
\bottomrule
\end{NiceTabular}
\end{table*}

\begin{table*}[t]
\centering
\caption{Comparison between our pipeline and existing methods (cf. \cref{sec:experimental_setup:sota}), depending on the input data. Results correspond to the AUROC's mean ± standard deviation over the test sets in a 5-fold cross-validation.}
\centering
\label{tab:sota_comparison}
\begin{NiceTabular}{cc ccc}
\toprule
\makecell{Tabular \\ descriptors} & Imaging data & \makecell{MMCL~\cite{hager_best_2023} \\ (\textasciitilde 36.7M param.)} & \makecell{IRENE~\cite{zhou_transformer-based_2023} \\ (\textasciitilde 104M param.)} & \makecell{Ours \\ (\textasciitilde 1.12M param.)} \\
\cmidrule(lr){1-2}
    \cmidrule(lr){3-5}
\Block{2-1}{all} & A4C/A2C ED/ES & 56.5 $\pm$ 6.3 & -- & --\\
    & Time-series descriptors & -- & 81.3 $\pm$ 11.1 & \underline{95.3 $\pm$ 3.1} \\
\cmidrule(lr){1-5}
\Block{2-1}{\makecell{clinical \\ only}} & A4C/A2C ED/ES & 56.0 $\pm$ 9.0 & -- & -- \\
    & Time-series descriptors & -- & 85.9 $\pm$ 15.2 & \textbf{96.8 $\pm$ 3.1} \\
\bottomrule
\end{NiceTabular}
\end{table*}

\begin{table}[t]
\centering
\caption{Ablation study of configurations of time-series tokenizer and multimodal attention for fusing heterogeneous data. Inputs are \textit{clinical only} tabular descriptors and time-series descriptors. Results correspond to the mean ± standard deviation over the test sets in a 5-fold cross-validation.}
\label{tab:ablation_multimodal_fusion}
\begin{NiceTabular}{cc c ccc c}
\toprule
\Block{3-1}{\rotate \makecell{Time-series \\ tokenizer}} & \Block{3-1}{\makecell{Bidirectional \\ attention \\ blocks}} & \Block{1-5}{XTab w/ random weights init.} & & & & \\
\cmidrule(lr){3-7}
    & & \Block{2-1}{AUROC} & \Block{1-3}{Per-class accuracy} & & & \Block{2-1}{\makecell{\# \\ param.}} \\
\cmidrule(lr){4-6}
    & & & wht & ctrl & unctrl & \\
\cmidrule(lr){1-2}
    \cmidrule(lr){3-7}
\Block{2-1}{\rotate linear}
    & \xmark & \res{93.0}{4.7} & \res{70.0}{44.7} & \res{88.5}{6.3} & \res{65.5}{25.0} & 891K \\
    & \cmark & \res{91.3}{4.2} & \res{33.0}{32.7} & \res{89.2}{6.6} & \textbf{\res{82.0}{10.4}} & 2\,079K \\
\cmidrule(lr){1-7}
\Block{2-1}{\rotate Transformer}
    & \xmark & \textbf{\res{96.8}{3.1}} & \textbf{\res{90.0}{13.6}} & \textbf{\res{94.2}{6.4}} & \res{78.3}{27.3} & 1\,115K \\
    & \cmark & \underline{\res{95.8}{4.4}} & \underline{\res{80.0}{27.3}} & \underline{\res{92.8}{5.0}} & \underline{\res{82.0}{24.0}} & 2\,303K \\
\bottomrule
\end{NiceTabular}
\end{table}

\begin{table}[t]
\centering
\caption{Ablation study of configurations of output token representation and classification formulation to structure the population representation. Inputs are \textit{clinical only} tabular descriptors and time-series descriptors. Results correspond to the mean ± standard deviation over the test sets in a 5-fold cross-validation. \# parameters are omitted because they do not change significantly.}
\label{tab:ablation_latent_representation}
\begin{NiceTabular}{cc c ccc}
\toprule
\Block{3-1}{\makecell{Token \\ representation}} & \Block{3-1}{\makecell{Ordinal \\ constraint}} & \Block{1-4}{XTab w/ random weights init.} & & & \\
\cmidrule(lr){3-6}
    & & \Block{2-1}{AUROC} & \Block{1-3}{Per-class accuracy} & & \\
\cmidrule(lr){4-6}
    & & & wht & ctrl & unctrl \\
\cmidrule(lr){1-2}
    \cmidrule(lr){3-6}
\Block{2-1}{\makecell{Sequence \\ pooling}}
    & \xmark & \underline{\res{97.9}{2.0}} & \underline{\res{85.0}{22.3}} & \underline{\res{93.5}{6.3}} & \underline{\res{83.4}{16.8}} \\
    & \cmark & \res{93.8}{5.1} & \res{70.0}{32.5} & \res{92.1}{8.1} & \res{60.3}{22.2} \\
\cmidrule(lr){1-6}
\Block{2-1}{\cls}
    & \xmark & \textbf{\res{98.0}{2.0}} & \underline{\res{85.0}{22.3}} & \res{92.8}{5.0} & \textbf{\res{84.6}{14.4}} \\
    & \cmark & \res{96.8}{3.1} & \textbf{\res{90.0}{13.6}} & \textbf{\res{94.2}{6.4}} & \res{78.3}{27.3} \\
\bottomrule
\end{NiceTabular}
\end{table}

\subsubsection{XTab Architecture Trained from Scratch Outperforms Other Models}
\label{sec:results:multimodal_fusion:xtab_architecture}
The design choices that most impact classification performance are the specific architecture configuration and how the weights are initialized. \Cref{tab:ablation_data_architecture_init} compares two models implementing the FT-Transformer architecture (cf. \cref{tab:fttransformer_configurations}). \textit{Tiny FT-Transformer} is the optimal configuration with very few parameters, while \textit{XTab} is the configuration used by the foundation model~\cite{zhu_xtab_2023}. For XTab, we tested two weights initialization: one with the foundation model's published weights, and one with the same architecture but randomly initialized weights (\textit{random}).

XTab performs better than the tiny model, up to 6 points more in AUROC and with a standard deviation (SD) reduced by up to 4.2 points depending on the configuration. This shows that the XTab architecture is able to generalize well and avoids significantly overfitting, especially considering the similar size between the output features (192D) and the number of training/validation samples (171/20) in our dataset.

However, the foundation model's pretrained weights gave surprising results. The model performs better with randomly initialized weights than with XTab's published weights, improving AUROC between 3.0 and 7.1 points depending on the input data. SD is also reduced on average by 1.6 points. This suggests that the foundation weights might not generalize well from the domains used for training~\cite{zhu_xtab_2023}, and that the architecture itself is more important.

Given these results, the following subsections and \cref{tab:ablation_multimodal_fusion,tab:sota_comparison,tab:ablation_latent_representation} focus on XTab with randomly initialized weights.

\subsubsection{Clinical Data Combined with Time-series Image Descriptors Enables the Best Performance}
\label{sec:results:multimodal_fusion:data}
To study the role of the data provided to the model, \cref{tab:ablation_data_architecture_init} reports performance over different configurations of tabular (cf. \cref{sec:experimental_setup:framework_config}) and time-series descriptors (cf. \cref{sec:experimental_setup:image}). \textit{Tabular descriptors} refer to the subsets of clinical data, while \textit{time-series descriptors} indicate whether unimodal tabular data was used (\xmark) or image-based data was included (\cmark).

Under the best model configuration, i.e. \textit{XTab} with \textit{random weights init.}, the best data configuration is the combination of time-series descriptors with clinical data excluding descriptors derived from echocardiograms (i.e. \textit{clinical only}). This makes sense if we consider that tabular descriptors not derived from images already provide a lot of information, evidenced by the unimodal model on \textit{clinical only} data reaching 95.8 AUROC. Time-series automatically extracted from echocardiograms are robust and fine-grained enough to provide complementary information to bring AUROC up to 96.8. This difference ends up not being statistically significant, although more decisive improvements might have been expected. However, this result can be explained by the added value from time-series’ rich information being less noticeable on high-level tasks, like classification, than on more detailed analysis (cf. \cref{sec:results:continuous_strat}). As for \textit{all} tabular descriptors derived from images, they add redundancy and noise, and their number (62 vs 28 \textit{clinical only}) complicates the fusion problem without significantly improving the results, even leading to a slight deterioration.

\subsubsection{Data Tokenization Leveraging Prior Knowledge Has More Impact Than Multimodal Fusion Strategy}
\label{sec:results:multimodal_fusion:tokenization}
To study our multimodal pipeline, we start by validating the relevance of working on time-series extracted from image rather than the images directly. \Cref{tab:sota_comparison} compares our model to MMCL~\cite{hager_best_2023}, a \sota\ model for combining image and tabular data (cf. \cref{sec:experimental_setup:sota:bob}). MMCL's results fall drastically below ours by around 40 points in AUROC. This confirms that training multimodal models directly on images for small datasets like ours is inadequate, and that image preprocessing guided by prior knowledge of relevant information, i.e. time-series descriptors, is necessary.

The next component to analyze is the token fusion strategy. The reference \sota\ method in this case is IRENE~\cite{zhou_transformer-based_2023}, and its approach to cross-attention with bidirectional attention blocks. \Cref{tab:sota_comparison} also compares our model to IRENE, with its tokenization adapted to use time-series (cf. \cref{sec:experimental_setup:sota:irene}). Depending on the tabular subset, our model is better by 11-14 points in AUROC. We explain IRENE's worse performance and higher SD by the fact that, with 104M parameters compared to our model's 1.1M, it is overparameterized for the task, and consequently less stable during training.

Although the full IRENE model underperformed, this alone does not invalidate their bidirectional attention block. Therefore, we integrated just this block in our framework (cf. \cref{fig:transformer-encoder}) and report the results in \cref{tab:ablation_multimodal_fusion}. \Cref{tab:ablation_multimodal_fusion} also validates another component related to token processing, by comparing two configurations of time-series tokenization: a simple \textit{linear} projection and the \textit{time-series Transformer} described in \cref{sec:method:tokenization:time-series}. When configuration changes are tweaks on top of the best combination of data and backbone architecture, differences in AUROC are getting smaller. Therefore, we add per-class accuracy scores to get a more comprehensive picture of the models' performance. However, accuracies have high SD on the minority classes (\textit{wht} and \textit{unctrl}). This is explained by the class imbalance and dataset partitions, which combined mean that there are only around 4 \textit{wht} and 15 \textit{unctrl} patients in each test set. For these classes, misclassifying a single sample can cause up to a 25\% drop in accuracy.

Results show that time-series tokenization has more impact than multimodal attention. The Transformer improves results by 3.8-4.5 points in AUROC and up to 47 points in accuracy. Results are also more stable for minority classes, compared to the poor per-class accuracies (33 and 65.5\%) and high SD (above 30) with the linear tokenizer. These observations confirm our hypothesis that time-series contain complex patterns which benefit from a more expressive tokenizer.

As for bidirectional attention, its impact is mostly negative. The difference is not as large as between time-series tokenizers, but AUROC is worse by 1.0-1.7 points, and per-class accuracies are less stable, falling as low as 33\%. We do not interpret these results as contradicting the original paper's findings, but rather to mean that the increase in parameters, which more than doubles the model's size, becomes too much for datasets like ours and leads to worse results.

\subsubsection{Best Representation Is Obtained by \cls\ Token}
\label{sec:results:multimodal_fusion:representation}
The final component left to discuss is how output features are represented and constrained. Regarding the definition of output features, \cref{tab:ablation_latent_representation} compares the standard \cls\ token with sequence pooling~\cite{hassani_escaping_2022}, which computes a dynamic average of all output tokens, similar to the fixed averaging done by IRENE~\cite{zhou_transformer-based_2023}. The impact of the ordinal constraint on both output representations is also reported.

The \cls\ token provides comparable results to sequence pooling when using a cross-entropy loss with no ordinal constraint (\xmark). However, adding the ordinal constraint (\cmark) degrades sequence pooling performance by 4.1 points in AUROC and 15 to 23 points in accuracy on minority classes. In contrast, the impact is mixed on the \cls\ token, with a slight decrease in AUROC and \textit{unctrl} accuracy, balanced by a comparable increase in \textit{wht} and \textit{ctrl} accuracies. Regarding the ordinal constraint, Beckham and Pal explained that their formulation regularizes the predicted probabilities, which might not help top-1 accuracy but could improve scores based on probabilities over multiple labels. They also noted the importance of learning a good temperature parameter $\tau$ (cf. \cref{fig:ordinal_classification}) to reach results on par with cross-entropy, while recognizing that training could be less stable~\cite{beckham_unimodal_2017}. We attribute our mixed results to this dual effect between regularizing the probabilities and a less stable training. In the end, the use of the ordinal formulation remains justified because it enables the continuous representation of patients, which we study in detail in the next subsection.

\subsection{Continuous Stratification Interpretability}
\label{sec:results:continuous_strat}
As mentioned earlier, we optimized our framework's classification scores assuming that better classification leads to representations that better characterize hypertension. We have observed empirically that because of hypertension severity’s inherent continuity, the ordinal classification constraint incentivizes smooth transitions in the latent space rather than hard clusters. In this context, better classification can be expected to correlate to a more informative manifold.

In the next subsections, we study the continuous latent representation learned by our model to see if it can reveal new, clinically interpretable markers of hypertension. Since there were no continuous targets to directly supervise this task, training samples are not biased in this evaluation compared to test samples. Therefore, following experiments use the full dataset to get as dense and representative of a distribution as possible.

Furthermore, the ablation studies discussed in \cref{sec:results:multimodal_fusion} allowed us to identify an optimal configuration (in bold in \cref{tab:ablation_data_architecture_init,tab:sota_comparison,tab:ablation_multimodal_fusion,tab:ablation_latent_representation}). Since we trained each configuration on 5 different folds, but only have the space to study the representation of one of them, we selected the model with the most uniformly distributed patients across the continuum. We defined a uniform distribution of patients as having low variability in the number of patients per discrete interval, i.e. bin, of the continuum.

\subsubsection{Stratification Fuzziness Correlates to Model Variability}
\label{sec:results:continuous_strat:reproducibility}
We started our analysis by computing the smallest mean absolute error (MAE) between the models' predicted $p$ for each patient and the average $p$ across the 5 models trained on different folds. Our chosen model's MAE is only 5.7\%. Given that, to the best of our knowledge, no previous work on classifying/phenotyping cardiovascular populations studied the robustness of their representation across trainings and partitionings~\cite{shah_phenomapping_2015,zheng_pathological_2020,loncaric_automated_2021}, we consider this result significant.

\begin{figure}[t]
    \includegraphics[width=0.95\columnwidth]{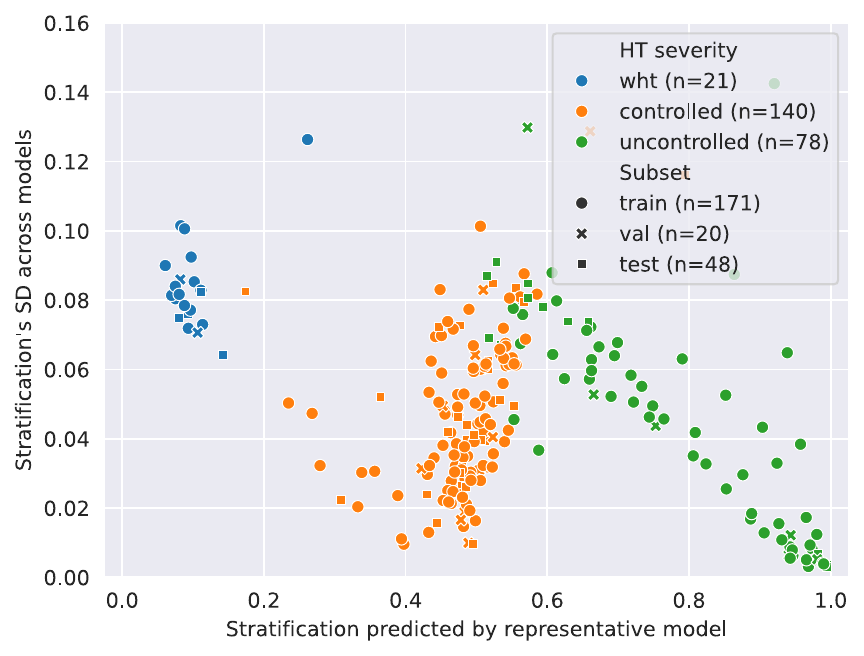}
    \caption{Visualization of the continuous stratification's standard deviation (SD) across models trained on different partitionings of the data w.r.t. the predictions of one model, selected as the most representative. The standard deviation tends to increase near the frontiers between labels, especially between controlled and uncontrolled. This means that patients considered ambiguous by the chosen model are indeed stratified with more variability across the ensemble of models.}
    \label{fig:continuum_variability}
\end{figure}

To understand how the variability between models manifests itself on a per patient basis, we look at the standard deviation (SD) between models' predicted stratifications. \Cref{fig:continuum_variability} plots this SD on each patient w.r.t. the stratification predicted by the representative model. This highlights that the variability is not constant along the stratification domain, i.e. a uniform band. The variability is higher in the transition zones, i.e. around 0.2 and 0.6 on the x-axis, and gradually decreases on both sides of these transitions. This increase in SD is much more striking in the transition from \textit{controlled} and \textit{uncontrolled} patients than from \textit{wht} to \textit{controlled} patients. We attribute this to the models' higher variability overall in their predictions for \textit{wht} patients. Indeed, the SD for \textit{wht} patients is between 0.06-0.10, whereas it can go as low as 0-0.01 for \textit{controlled} or \textit{uncontrolled} patients. This is probably due to the fact that there are relatively few \textit{wht} patients (n=21).

This result is desirable, because it shows that when the representative model's stratification suggests fuzzy associations to target labels, it correlates with a higher variability in stratification across the set of models. This indicates that the model's fuzziness is representative of the framework's variability. It is further proof that the representation learned by our model achieved the stated goal of stratifying the patients beyond the provided target labels.

\begin{figure*}[thb]
\begin{minipage}{0.64\linewidth}
    \begin{subfigure}[t]{0.49\textwidth}
        \begin{center}
        \includegraphics[width=\textwidth]{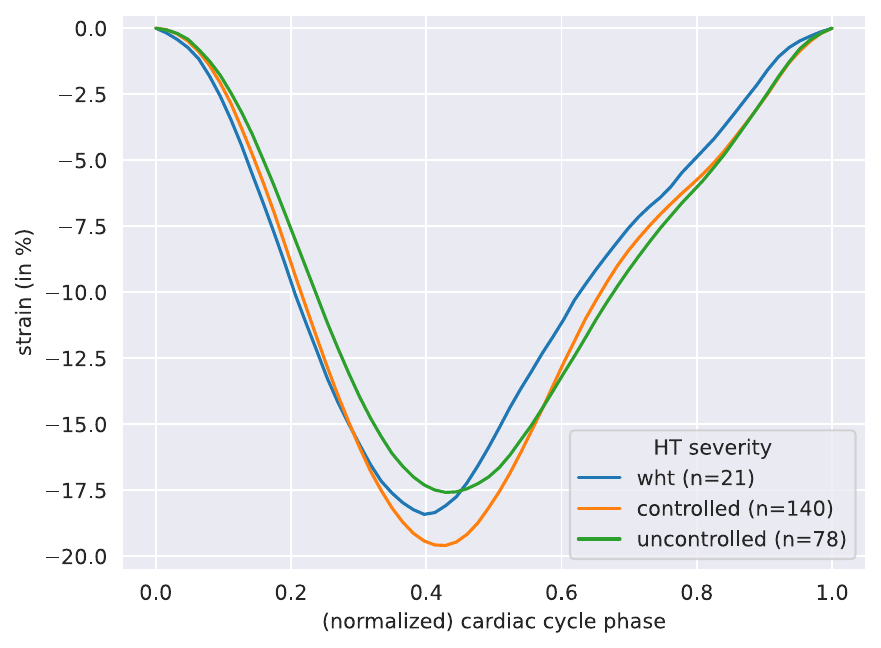}
        \end{center}
        \caption{Apical 4 Chamber (A4C) Global Longitudinal Strain (GLS) w.r.t. target labels}
        \label{fig:gls_wrt_ht_severity}
    \end{subfigure}
    \begin{subfigure}[t]{0.49\textwidth}
        \begin{center}
        \includegraphics[width=\textwidth]{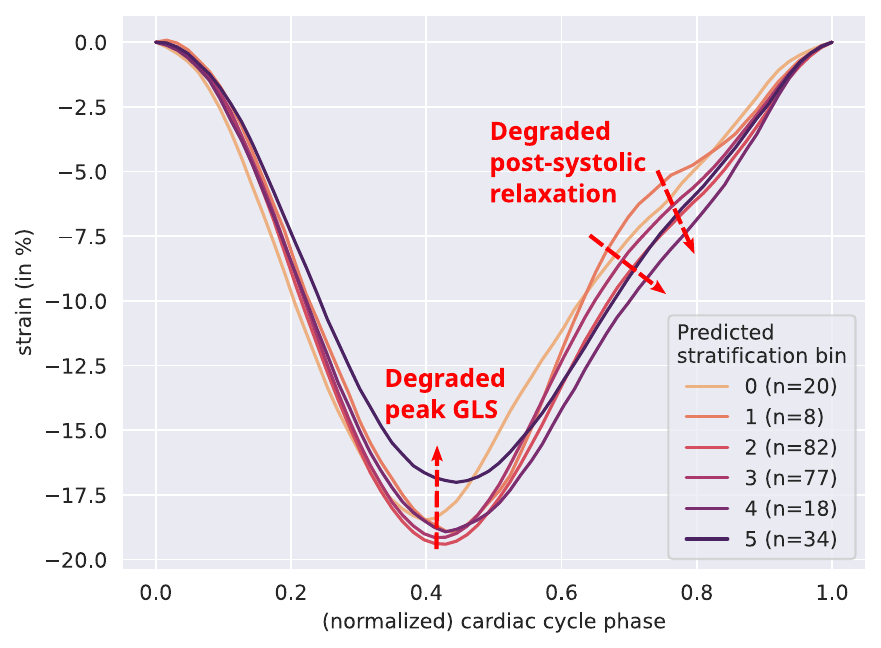}
        \end{center}
        \caption{A4C GLS w.r.t. stratification prediction}
        \label{fig:gls_wrt_continuum_bins}
    \end{subfigure}
    \begin{subfigure}[t]{0.49\textwidth}
        \begin{center}
        \includegraphics[width=\textwidth]{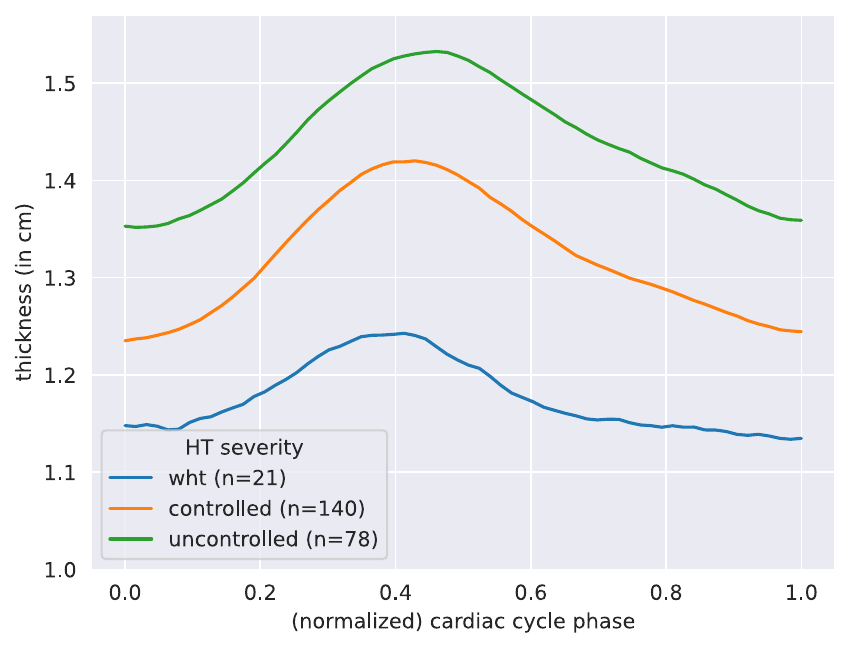}
        \end{center}
        \caption{A4C Basal Septal thickness (BST) w.r.t. target labels}
        \label{fig:bst_wrt_ht_severity}
    \end{subfigure}
    \begin{subfigure}[t]{0.49\textwidth}
        \begin{center}
        \includegraphics[width=\textwidth]{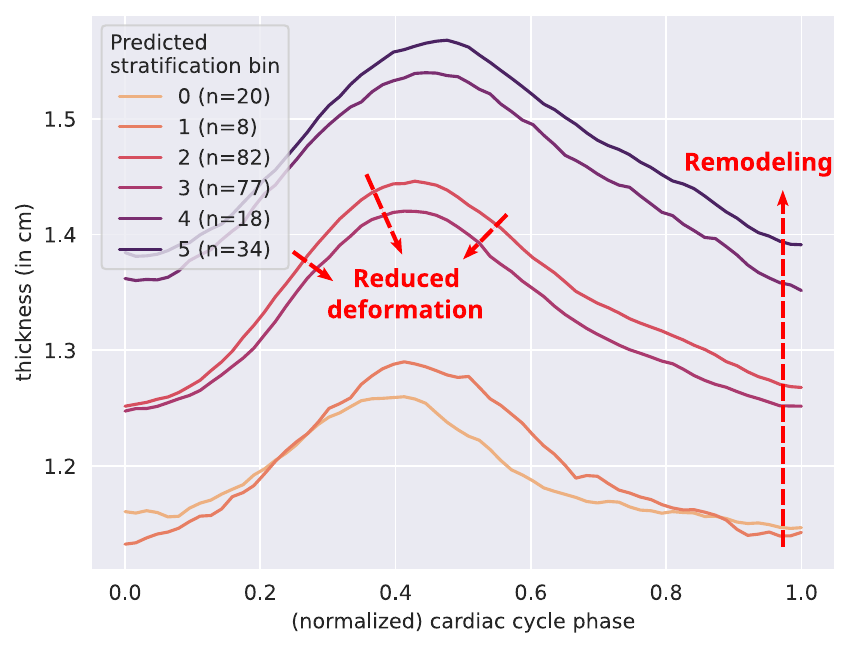}
        \end{center}
        \caption{A4C BST w.r.t. stratification prediction}
        \label{fig:bst_wrt_continuum_bins}
    \end{subfigure}
\end{minipage}
\begin{minipage}{0.35\linewidth}
    \begin{subfigure}[t]{\textwidth}
        \begin{center}
        \includegraphics[width=\textwidth]{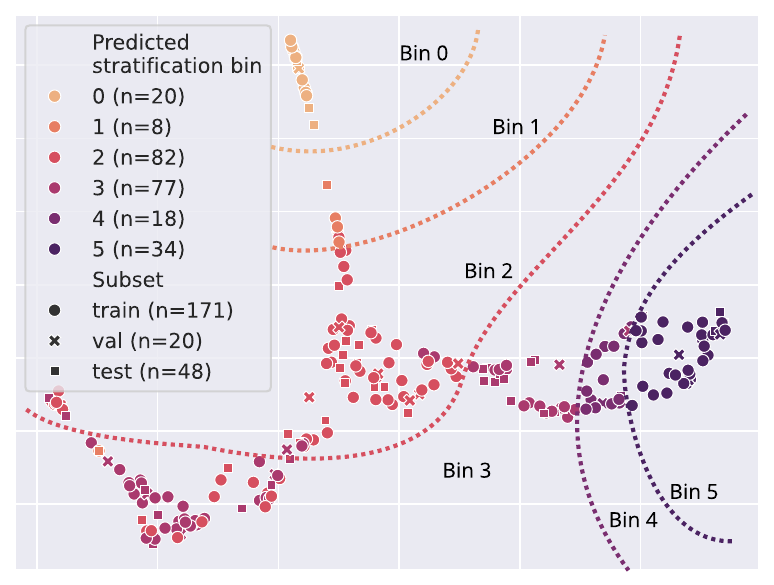}
        \end{center}
        \caption{Embedding using PaCMAP (cf. \cref{fig:embeddings}) colored w.r.t. stratification prediction bins. The bins are in ascending order of HT severity. They were obtained by dividing the stratification $\in [0,1]$ into 6 equal intervals. Note that PaCMAP's non-linearity complicates the visualization of the bins. Thus, while there exists a linear direction along which the bins are strictly ordered in the 192-dimensional feature space, some bins overlap each other in the 2D embedding, e.g. in the lower left of the plot.}
        \label{fig:bins_embedding}
    \end{subfigure}
\end{minipage}
    \caption{Illustration of the repercussions of HT severity on cardiac function descriptors. Each plot shows the average curve for one cardiac function descriptor by groups of patients (whose sizes are specified in the legends). The patient groups are determined either based on the reference HT severity labels (a, c), or on discrete bins derived from our continuous stratification (b, d). \subref{fig:bins_embedding} corresponds to the same 2D embedding as in \cref{fig:embeddings}, but w.r.t. these stratification bins. Both the target labels and stratification bins display multiple phenotypes, but the continuous stratification describes in more incremental steps the evolution from normotensive cases (\textit{wht}/\textit{0}) to severe HT (\textit{uncontrolled}/\textit{5}).}
    \label{fig:time_series_attrs_wrt_stratification}
\end{figure*}

\begin{figure}[thb]
    \begin{subfigure}[c]{0.49\columnwidth}
        \begin{subfigure}[t]{\textwidth}
            \begin{center}
            \includegraphics[width=\textwidth]{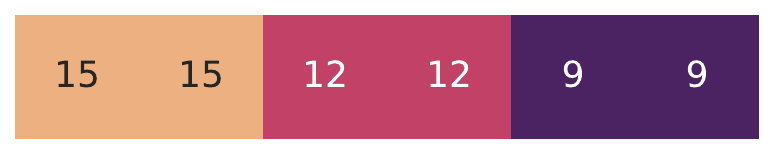}
            \caption{Lateral e$^\prime$}
            \end{center}
            \label{fig:lateral_e_prime_wrt_continuum_bins}
        \end{subfigure}
        \begin{subfigure}[t]{\textwidth}
            \begin{center}
            \includegraphics[width=\textwidth]{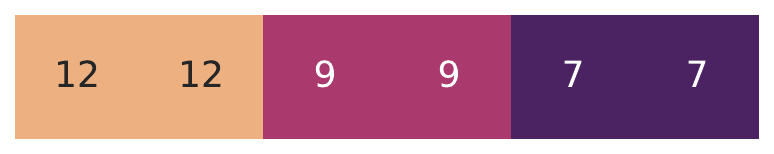}
            \caption{Septal e$^\prime$}
            \end{center}
            \label{fig:septal_e_prime_wrt_continuum_bins}
        \end{subfigure}
        \begin{subfigure}[t]{\textwidth}
            \begin{center}
            \includegraphics[width=\textwidth]{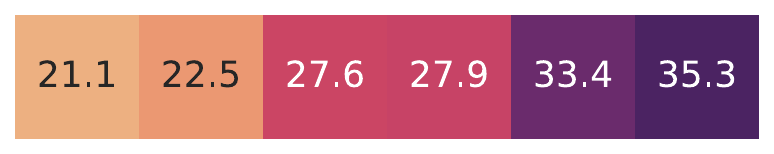}
            \caption{Left atrium volume}
            \end{center}
            \label{fig:la_volume_wrt_continuum_bins}
        \end{subfigure}
        \caption*{Linearly correlated to HT severity}
    \end{subfigure}
    \begin{subfigure}[c]{0.49\columnwidth}
        \centering
        \begin{subfigure}[t]{\textwidth}
            \begin{center}
            \includegraphics[width=\textwidth]{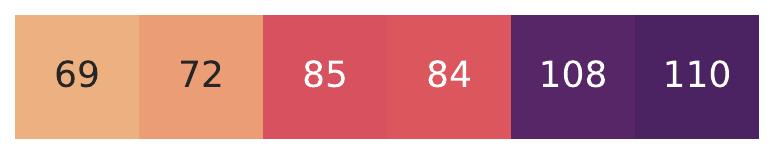}
            \caption{Indexed LV mass}
            \end{center}
            \label{fig:lvm_ind_wrt_continuum_bins}
        \end{subfigure}
        \begin{subfigure}[t]{\textwidth}
            \begin{center}
            \includegraphics[width=\textwidth]{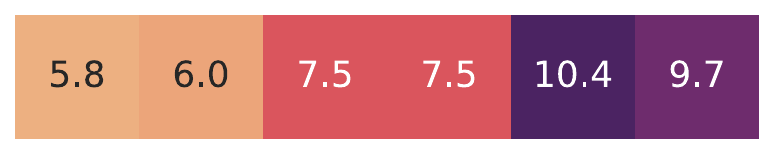}
            \caption{Ratio of e over e$^\prime$}
            \end{center}
            \label{fig:e_e_prime_ratio_wrt_continuum_bins}
        \end{subfigure}
        \caption*{Non-linearly correlated to HT severity}
    \end{subfigure}
    \caption{Heatmaps of scalar cardiac function descriptors known to be linked to HT, averaged over each of the 6 bins derived from our continuous stratification (cf. \cref{fig:bins_embedding}), in ascending order. Colors show the progression from normal values (light) to pathological values (dark).}
    \label{fig:tabular_attrs_wrt_continuum_bins}
\end{figure}

\subsubsection{Continuum Highlights New Prospective Markers of Hypertension in Time-series}
\label{sec:results:continuous_strat:time_series_markers}
\Cref{fig:time_series_attrs_wrt_stratification} studies patterns on two time-series descriptors with respect to both the target labels and continuous stratification, Global Longitudinal Strain (GLS) and Basal Septal Thickness (BST), which are important parameters of left ventricular function and remodeling~\cite{salte_artificial_2021,marciniak_septal_2021}.

GLS measures the deformation of the myocardium as the [negative] ratio of the contraction along its contour w.r.t. when the ventricle is most dilated at ED. Cardiologists often use the smallest GLS value, the peak GLS, as a biomarker of systolic function, along with ejection fraction (EF), i.e. the ratio of the variation of the volume of the ventricle. However, using the full GLS curve, instead of the scalar peak GLS and EF, can provide a more detailed analysis of the specific mechanisms affected by HT~\cite{cikes_role_2010}. BST presents a similar story. It measures the thickness of the myocardium at the base of the septum, i.e. the wall between the left and right ventricles. The septum can become thicker in patients suffering from HT, as a response to the increased pressure (afterload) from the LV side~\cite{marciniak_septal_2021}. BST is measured statically from parasternal acquisitions in typical clinical workflows, but studying its variation over the cardiac cycle could provide additional insights.

\Cref{fig:gls_wrt_ht_severity,fig:gls_wrt_continuum_bins} present average GLS curves over target HT severity labels and a discretized version of the continuous stratification from our method (cf. \cref{fig:bins_embedding}). Two markers related to HT severity can be observed in these curves. The first one is that peak GLS actually improves in controlled patients compared to normotensive subjects (\textit{wht}), likely due to their hypertensive treatment. Peak GLS then degrades for uncontrolled patients, once their treatment no longer manages the effects of HT. This first marker is observable using both the target labels (\cref{fig:gls_wrt_ht_severity}) and the continuous stratification (\cref{fig:gls_wrt_continuum_bins}). However, only the continuous stratification highlights the second marker, namely the degradation of post-systolic relaxation in intermediate to early severe cases (bins 1-4). Post-systolic relaxation is measured through the slope of the strain curve after systole, i.e. between 0.4 and 0.8 on the x-axis, with healthier subjects presenting earlier, sharper slopes. Thus, while bins 1-4 present similar peak GLS, post-systolic relaxation gradually happens later and slower. Thus, thanks to our continuous stratification approach, we highlighted that post-systolic relaxation could be an early marker of worsening HT, before meaningful changes in peak GLS.

\Cref{fig:bst_wrt_ht_severity,fig:bst_wrt_continuum_bins} show similar curves, but for septal thickness. Again, both the target labels and continuous stratification exhibit similar tendencies, with BST being overall correlated to HT severity. However, the granularity on intermediate cases (bins 2-3) suggests the relation between BST and HT severity might not be linear. The septum of patients in bin 2 thickens more during end-systole, around 0.4 on the x-axis, than that of patients in bin 3. In other terms, on the CARDINAL cohort, HT first leads to a less deformable septum, before the septum thickens to accommodate the higher pressure. These results confirm the established notion that HT leads to septal remodeling~\cite{marciniak_septal_2021}, while suggesting that altered deformation patterns could also be markers of HT progression.

\subsubsection{Continuum Reranks Importance of Tabular Markers of Hypertension}
\label{sec:results:continuous_strat:tabular_markers}
Finally, \cref{fig:tabular_attrs_wrt_continuum_bins} presents an analogous study on scalar cardiac function descriptors. An expert cardiologist analyzed the tabular descriptors, aggregated with respect to bins on our continuous stratification (cf. \cref{fig:bins_embedding}), to see if relevant tendencies existed across the bins. Five parameters linked to HT~\cite{mancia_2023_2023} were of particular interest: lateral and septal mitral annulus velocity, commonly referred to as \textit{e$^\prime$} (a, b), left atrium volume \subref{fig:la_volume_wrt_continuum_bins}, indexed LV mass \subref{fig:lvm_ind_wrt_continuum_bins}, and the ratio of \textit{e} over \textit{e$^\prime$} \subref{fig:e_e_prime_ratio_wrt_continuum_bins}. While the indexed LV mass and ratio of \textit{e} over \textit{e$^\prime$} are usually the measures most relied upon, \cref{fig:tabular_attrs_wrt_continuum_bins} illustrates that they do not always progress linearly with HT severity, unlike the other descriptors. Their locally non-linear relationships could not be identified from the target HT severity labels because they group too many patients together. Therefore, our continuous stratification paves the way for renewed investigations into the predictive power of known scalar markers.

\section{Discussion and Conclusion}
\label{sec:discussion&conclusion}
In this work, we proposed a framework for fusing multimodal tabular and 2D+time echocardiograms to learn a continuous stratification of patients, given limited data with categorical labels. The framework is designed to target difficult-to-characterize pathologies by, on one hand, combining complementary descriptors derived from images and EHRs, and, on the other hand, projecting the patients along an interpretable pathological continuum.

To extract relevant cardiac function descriptors from the echocardiograms, our pipeline relies on accurate and reproducible segmentations of the left ventricle and myocardium. To achieve such segmentations, we use a state-of-the-art model which has shown accuracy within intra-observer variability, validated for anatomical and temporal coherence (cf. \cref{sec:experimental_setup:image}). This echocardiograms processing step serves the double role of preprocessing imaging data for the rest of the pipeline, and of extracting detailed cardiac function descriptors for later analysis.

To tackle the challenge of multimodal fusion, we built our approach around tabular Transformers. Inspired by mid-fusion methods, the scalar and time-series descriptors, extracted from imaging data, are tokenized and combined to tabular tokens. We showed that, in the context of our application, an expressive Transformer tokenizer that can leverage the strong inductive bias behind the time-series is preferable to generic multimodal fusion strategies. For our model's backbone, we relied on the architecture of the XTab foundation model for tabular data. In our limited data setting, it performed better than smaller configurations of the same architecture. Surprisingly, it even performed best when trained from scratch, rather than using the pretrained weights. Less surprisingly, it performed better than models that are bigger by an order of magnitude or two. These findings hint at XTab corresponding to an ideal compromise between expressivity and parametrization that generalizes well even on small datasets, which warrants further research. We also framed the task as an ordinal classification to integrate the order of the labels and induce a continuously interpretable stratification in the latent space, while keeping performance on par with a cross-entropy baseline. Finally, we showed that the continuous stratification is coherent with prior physiological knowledge, i.e. the progression in target labels’ phenotypes, while allowing for more detailed analysis of local variations within established phenotypes. Eventually, the continuous stratification could also be used to personalize hypertension prognosis~\cite{zhang_breaking_2025}. However, the robustness of the stratification at the patient level will need to be validated, notably with a rigorous estimation of the uncertainty of the predictions, before such claims can be verified.

Compared to previous works, our approach stands out by working with tabular and imaging data, and by predicting a precise stratification of patients. The work most directly comparable to ours is the IRENE multimodal Transformer~\cite{zhou_transformer-based_2023}, which fuses images and unstructured medical records to classify pulmonary diseases. However, their datasets of 2.4K and 44K+ samples allowed them to train a large model directly on images and unstructured medical records. In our case, the poor performance of MMCL~\cite{hager_best_2023}, a \sota\ method for combining imaging and tabular data, demonstrates that our dataset is too limited to work directly on images. Therefore, we implemented a pipeline leveraging prior expert knowledge to extract relevant descriptors from images and EHRs. On top of that, we also used a much smaller Transformer architecture to significantly outperform IRENE when trained on the same preprocessed data. Our pipeline demonstrated good performance considering the dataset's challenging small size and class imbalance. The selected configuration's classification accuracy on the minority class (\textit{wht}) is comparable to that of the dominant classes (\textit{controlled} / \textit{uncontrolled}), indicating that the model avoids biasing its representation towards more common phenotypes.

\subsection{Limitations \& Future Research}
\label{sec:discussion&conclusion:limitations}

\subsubsection{Dataset}
\label{sec:discussion&conclusion:limitations:dataset}
One possible criticism against our study are the limits of the CARDINAL dataset. Among competing \sota\ methods for multi-modal healthcare classifications, even the ones most similar to ours~\cite{zhou_transformer-based_2023,hager_best_2023,schilcher_fusion_2024} are trained on at least thousands of samples. We would argue that CARDINAL is representative of pilot clinical studies in cardiac imaging, paired with some unique advantages. First, its size is in line with the scale of other targeted clinical cohorts~\cite{loncaric_automated_2021}, and only one order of magnitude smaller than the 2.4K samples IRENE was trained on. Second, the data was collected across two hospitals, although the limited number of subjects prevented us from holding out a full site for testing. Third, CARDINAL aims to provide representative samples from all over the HT continuum. By contrast, the larger datasets used by competing methods are aimed at multi-disease classification, with no notion of disease progression. Given our stated goal of leveraging rich data to provide a more detailed stratification, the use of the CARDINAL dataset is justified over other, potentially larger, datasets. It follows that a pipeline like ours, which can outperform its closest counterparts and achieve excellent results on datasets like CARDINAL, is highly relevant to enable pilot clinical studies to use SOTA deep learning-based methods. In the end, potential markers of hypertension highlighted in this study will have to be validated on larger, even more diverse datasets. However, this further step towards direct clinical impact is out of the scope of the current paper.

\subsubsection{Failure cases}
\label{sec:discussion&conclusion:limitations:failure}
We also looked at cases that our selected model (cf. \cref{sec:results:continuous_strat}) misclassified to observe if it failed more on specific types of cases. The model achieved 100\% test accuracy on the minority \textit{wht} class, only misclassifying the dominant \textit{controlled} and \textit{uncontrolled} classes, further reinforcing the observation that it performs well even on the minority class. Looking at the continuous stratification, the average distance between misclassified predictions and their true label’s nearest boundary in the continuum is 0.12. This means that errors typically correspond to borderline cases being predicted slightly off in the continuum, passed the hard label threshold. Finally, an expert cardiologist qualitatively reviewed representative misclassified samples. In five out of six cases, extreme parameters (e.g. a very young patient, exceedingly high doses of hypertensive treatments, etc.) could explain, if not justify, the model's predictions. Handling these rare outliers remains an open challenge for deep learning models, which future works could study to make the proposed method more robust and reliable for personalized prognosis.

\subsubsection{Uncertainty Estimation}
\label{sec:discussion&conclusion:limitations:uncertainty}
While not a focus of this paper, the ordinal classification formulation opens new potential estimators of uncertainty. We showed in \cref{fig:continuum_variability} that the stratification continuum between classes  correlates to higher inter-model variability. Future work could look at the correlation between the position on the continuum and established uncertainty estimation methods. In the same vein, per-sample softmax temperature $\tau$ (cf. \cref{fig:ordinal_classification}) can increase (decrease, respectively) the prediction's entropy by uniformizing (sharpening) the categorical distribution, making it a promising uncertainty measure.

\subsection{Conclusion}
\label{sec:discussion&conclusion:conclusion}
We presented a framework that learns a stratification of hypertensive patients by fusing 2D+time echocardiographic images with structured medical records data. We improved the relevance of the representation with respect to hypertension by combining rich time-series descriptors extracted from automatic segmentations with complementary descriptors from electronical health records. We showed that to leverage these time-series descriptors, the best approach in limited data settings is to use a dedicated Transformer for tokenization, rather than rely on parameter-heavy multimodal fusion schemes. For the Transformer backbone, the XTab foundation model's configuration provided the best compromise between expressivity and parameter count, but surprisingly performed better when trained from scratch than when using the foundation model's pretrained weights. Finally, we obtained a better joint representation by using the traditional \cls\ token than by averaging output tokens using sequence pooling. We also tailored this representation to difficult-to-characterize pathologies by using ordinal classification to learn a continuous stratification.

Additionally, we showed that our continuous stratification can be used as a tool to analyze patient populations, to gain better insights into how hypertension impacts clinically-relevant cardiac shape and motion descriptors. In contrast, clinicians' discrete stratification reference might miss or mischaracterize alterations of these descriptors. In the end, we hope these insights inspire further exploratory studies to characterize full pathological continuums, rather than discrete stages, from all the multimodal data available in healthcare settings.



\section*{Acknowledgment}
The authors thank Elisa Le Maout (ARC, Hôpital Lyon Sud, Hospices Civils de Lyon, Lyon, France) for her help with data collection for the CARDINAL dataset.

\bibliographystyle{IEEEtran}
\bibliography{clean-copy-TUFFC-14012-2025.R1}

\end{document}


\setcounter{table}{5}   

\begin{table*}[t]
\centering
\caption{List of the 46 patient descriptors for the CARDINAL dataset extracted from Electronic Health Records (EHRs). The descriptors are categorized by the exams/records from which they are obtained. In this case, the descriptors coming from images, i.e. TTE, were manually measured as part of the clinical workflow.}
\makegapedcells
\label{tab:records_descriptors}
\begin{NiceTabular}{l l l l}
\CodeBefore
 \rowlistcolors{2}{SeabornGray,white}[restart,cols={2-4}]
\Body
\toprule
Source &
    \mkcl{Abbreviation \\ (in code/figures)} &
    Unit/labels &
    Description \\
\midrule
\Block{3-1}{\rotate \makecell{General \\ info}}
    & age & years & Age \\
    & sex & M/W & Sex \\
    & bmi & kg/m\textsuperscript{2} & Body Mass Index (BMI) \\
\midrule
\Block{7-1}{\rotate Medical history}
    & hf & yes/no & Heart Failure \\
    & cad & yes/no & Coronary Artery Disease (CAD) \\
    & pad & yes/no & Peripheral Artery Disease (PAD) \\
    & stroke & yes/no & Stroke \\
    & tobacco & none/ceased/active & Tobacco \\
    & diabetes & yes/no & Diabetes \\
    & dyslipidemia & yes/no & Dyslipidemia \\
\midrule
\Block{3-1}{\rotate \makecell{Holter \\ monitor}}
    & sbp\_24 & mmHg & Systolic Blood Pressure (SBP) averaged over 24h \\
    & dbp\_24 & mmHg & Diastolic Blood Pressure (DBP) averaged over 24h \\
    & pp\_24 & mmHg & Pulse Pressure (PP) averaged over 24h \\
\midrule
\Block{3-1}{\rotate \makecell{Biological \\ exam}}
    & creat & \textmu mol/L & Plasma CREATinine level \\
    & gfr & mL/min/1.73m\textsuperscript{2} & Glomerular Filtration Rate (GFR) indexed to standard body surface area (1.73m\textsuperscript{2}) \\
    & nt\_probnp & pg/mL & Molar ratio of NT-proBNP \\
\midrule
\Block{10-1}{\rotate Treatment}
    & ddd & -- & Defined Daily Dose (DDD) of blood pressure medication \\
    & bradycardic & yes/no & Medication includes bradycardic agents (beta-blockers, diltiazem or verapamil) \\
    & ace\_inhibitor & yes/no & Medication includes Angiotensin-Converting Enzyme (ACE) inhibitors \\
    & arb & yes/no &  Medication includes Angiotensin Receptor Blockers (ARB) \\
    & tz\_diuretic & yes/no &  Medication includes ThiaZide (TZ) diuretics \\
    & central\_acting & yes/no &  Medication includes central-acting agents \\
    & beta\_blocker & yes/no &  Medication includes beta blockers \\
    & spironolactone & yes/no &  Medication includes spironolactone \\
    & alpha\_blocker & yes/no &  Medication includes alpha blockers \\
    & ccb & yes/no &  Medication includes Calcium Channel Blockers (CCB) \\
\midrule
\Block{20-1}{\rotate Transthoracic echocardiogram (TTE)}
     & sbp\_tte & mmHg & Systolic Blood Pressure (SBP) during TTE \\
     & dpb\_tte & mmHg & Diastolic Blood Pressure (SBP) during TTE \\
     & pp\_tte & mmHg & Pulse Pressure (SBP) during TTE \\
     & hr\_tte & Beats Per Minute (BPM) & Heart Rate (HR) during TTE \\
     & e\_velocity & m/s & E-wave (passive blood flow from left atrium to left ventricle) velocity\\
     & a\_velocity & m/s & A-wave (active blood flow caused by atrial contraction) velocity \\
     & mv\_dt & millisecond & Mitral Valve (MV) Decelaration Time (DT) \\
     & lateral\_e\_prime & cm/s & Lateral mitral annular velocity (e') \\
     & septal\_e\_prime & cm/s & Septal mitral annular velocity (e')\\
     & reduced\_e\_prime & yes/no & Reduced E': lateral e' $<$ 10 cm/s or septal e' $<$ 7 cm/s \\
     & e\_e\_prime\_ratio & -- & Ratio of E velocity over e': E/e' \\
     & d\_dysfunction\_e\_e\_prime\_ratio & yes/no & High ratio of E/e' indicating Diastolic dysfunction: E/e' $>$ 14\\
     & la\_volume & mL/m\textsuperscript{2} & Left Atrial (LA) volume indexed to body surface area (BSA) \\
     & dilated\_la & yes/no & \mkcl{Dilated LA volume indexed to BSA indicating diastolic dysfunction: \\ \textit{la\_volume} $>$ 34 mL/m\textsuperscript{2}} \\
     & ph\_vmax\_tr & yes/no & \mkcl{Pulmonary Hypertension (PH) indicated by peak Tricuspid Regurgitation (TR): \\ TR $>$ 2.8 m/s} \\
     & lvm\_ind & g/m\textsuperscript{2} & Left Ventricular Mass (LVM) indexed to BSA \\
     & lvh & yes/no & Left Ventricular Hypertrophy (LVH) \\
     & ivs\_d & cm & InterVentricular Septum (IVS) thickness at end-Diastole (D) \\
     & lvid\_d & cm & Left Ventricular Internal Diameter (LVID) at end-Diastole (D) \\
     & pw\_d & cm &  Left ventricular Posterior Wall (PW) thickness at end-Diastole (D) \\
\bottomrule
\end{NiceTabular}
\end{table*}

\begin{table*}[t]
\centering
\caption{List of the 7 patient descriptors for the CARDINAL dataset derived from a posterior assessment by a cardiologist.}
\makegapedcells
\label{tab:assessment_descriptors}
\begin{NiceTabular}{l l l l}
\CodeBefore
 \rowlistcolors{2}{SeabornGray,white}[restart,cols={2-4}]
\Body
\toprule
Source &
    \mkcl{Abbreviation \\ (in code/figures)} &
    Labels &
    Description \\
\midrule
\Block{7-1}{\rotate Hypertension assessment}
    & etiology & \mkcl{essential \\ secondary \\ pa} & \mkcl{Etiology of the HyperTension (HT): \\ \textit{essential}: HT without one distinct cause \\ \textit{secondary}: HT related to endocrine, renal or renovascular conditions \\ \textit{pa}: HT caused by Primary hyperAldosteronism (PA)} \\
    & nt\_probnp\_group & \mkcl{neutral \\ end\_organ\_damage \\ mortality\_rate} & \mkcl{Correlation with potential outcomes based on NT-proBNP rate: \\ \textit{neutral}: No known adverse effects \\ \textit{end\_organ\_damage}: Risk of at least one end/target organ impacted by HT \\ $\quad$ ($>$ 90 pg/mL for men, $>$ 142 pg/mL for women) \\ \textit{mortality\_rate}: Risk of mortality ($>$ 150 pg/mL)} \\
    & ht\_grade & 0--3 & \mkcl{Classification of HT based on measured Blood Pressure (BP) over 24h: \\ \textit{0 (normal)}: Systolic BP $<$ 130 mmHg and/or Diastolic BP $<$ 80 mmHg \\ \textit{Grade 1 HT}: 130-149 SBP and/or 80-89 DBP \\ \textit{Grade 2 HT}: 150-169 SBP and/or 90-99 DBP \\ \textit{Grade 3 HT}: SBP $\ge$ 170 and/or DBP $\ge$ 100} \\
    & ht\_severity & \mkcl{wht \\ controlled \\ uncontrolled} & \mkcl{Severity of HT manually determined by a cardiologist: \\ \textit{wht}: no positive diagnosis of HT (White coat HyperTension) \\ \textit{controlled}: under the recommended blood pressure level given the treatment \\ $\quad$ (grade 1 or lower) \\ \textit{uncontrolled}: above the recommended blood pressure level despite the treatment \\ $\quad$ (grade 2 or higher)} \\
    & diastolic\_dysfunction\_param\_sum & 0--4 & \mkcl{1 point per parameter of diastolic dysfunction (cf. \cref{tab:records_descriptors}): \\ \textit{dilated\_la}, \textit{reduced\_e\_prime}, \textit{d\_dysfunction\_e\_e\_prime\_ratio}, \textit{ph\_vmax\_tr}} \\
    & diastolic\_dysfunction & \mkcl{none \\ uncertain \\ certain} & \mkcl{Diagnosis of diastolic dysfunction, based on the sum of parameters: \\ \textit{none}: 0 or 1 parameter \\ \textit{uncertain}: 2 parameters \\ \textit{certain}: 3 or 4 parameters} \\
    & ht\_cm & \mkcl{none \\ uncertain \\ certain} & \mkcl{Diagnosis of HyperTensive CardioMyopathy (HT-CM) based on TTE analysis: \\ \textit{none}: no diastolic dysfunction and no Left Ventricular Hypertrophy (LVH) \\ \textit{uncertain}: uncertain diastolic dysfunction and no LVH \\ \textit{certain}: certain diastolic dysfunction and/or LVH} \\
\bottomrule
\end{NiceTabular}
\end{table*}

\begin{table*}[t]
\centering
\caption{List of the 25 patient descriptors extracted from segmentations of transthoracic echocardiogram (TTE) for the CARDINAL dataset. The descriptors are categorized by whether they are global biomarkers (\textit{Scalar}), or extracted frame-by-frame (\textit{Time-series}). In this case, all descriptors were extracted automatically from left ventricle and myocardium segmentations.}
\makegapedcells
\label{tab:segmentation_descriptors}
\begin{NiceTabular}{l l l l ccc l}
\CodeBefore
 \rowlistcolors{3}{SeabornGray,white}[restart,cols={2-8}]
\Body
\toprule
\Block{2-1}{Type} &
    \Block{2-1}{\mkcl{Abbreviation \\ (in code/figures)}} &
    \Block{2-1}{Unit} &
    \Block{2-1}{Count} &
    \Block{1-3}{Input views} & & &
    \Block{2-1}{Description} \\
\cmidrule(lr){5-7}
&
    &
    &
    &
    A4C & A2C & A4C+A2C &
    \\
\midrule
\Block{5-1}{\rotate Scalar}
    & edv & mL & 1 & & & \cmark & End-Diastole (ED) Volume (V) of the Left Ventricle (LV) \\
    & esv & mL & 1 & & & \cmark & End-Systole (ES) Volume of the LV \\
    & ef & \% & 1 & & & \cmark & Ejection Fraction (EF) of the LV \\
    & \mkcl{a4c\_ed\_sc\_[min$\mid$max] \\ a4c\_ed\_lc\_[min$\mid$max]} & dm\textsuperscript{-1} & 4 & \cmark & & & \mkcl{Min./max. myocardial curvature at the base of each wall in A4C ED: \\ \textit{A4C left wall}: septum (s) / \textit{A4C right wall}: lateral (l)} \\
    & \mkcl{a2c\_ed\_ic\_[min$\mid$max] \\ a2c\_ed\_ac\_[min$\mid$max]} & dm\textsuperscript{-1} & 4 & & \cmark & & \mkcl{Min./max. myocardial curvature at the base of each wall in A2C ED: \\ \textit{A2C left wall}: inferior (i) / \textit{A2C right wall}: anterior (a)} \\
\midrule
\Block{7-1}{\rotate Time-series}
    & lv\_area & cm\textsuperscript{2} & 2 & \cmark & \cmark & & Surface area of the LV \\
    & lv\_length & cm & 2 & \cmark & \cmark & & Distance between the LV's apex and midpoint at the base \\
    & gls & \% & 2 & \cmark & \cmark & & Global Longitudinal Strain (GLS) \\
    & ls\_left & \% & 2 & \cmark & \cmark & & \mkcl{Regional Longitudinal Strain (LS) at the base of the left wall \\ \textit{A4C left wall}: septum / \textit{A2C left wall}: inferior} \\
    & ls\_right & \% & 2 & \cmark & \cmark & & \mkcl{Regional Longitudinal Strain (LS) at the base of the right wall \\ \textit{A4C right wall}: lateral / \textit{A2C right wall}: anterior} \\
    & myo\_thickness\_left & cm & 2 & \cmark & \cmark & & \mkcl{Average myocardial thickness at the base of the left wall \\ \textit{A4C left wall}: septum / \textit{A2C left wall}: inferior} \\
    & myo\_thickness\_right & cm & 2 & \cmark & \cmark & & \mkcl{Average myocardial thickness at the base of the right wall \\ \textit{A4C right wall}: lateral / \textit{A2C right wall}: anterior} \\
\bottomrule
\end{NiceTabular}
\end{table*}